\pdfoutput=1

\documentclass[11pt]{article}

\usepackage[final]{acl}

\usepackage{times}
\usepackage{latexsym}
\usepackage{footmisc}

\usepackage[T1]{fontenc}

\usepackage[utf8]{inputenc}

\usepackage{microtype}

\usepackage{inconsolata}

\usepackage{graphicx}

\usepackage[utf8]{inputenc} 
\usepackage[T1]{fontenc}    
\usepackage{hyperref}       
\usepackage{url}            
\usepackage{booktabs}       
\usepackage{amsfonts}       
\usepackage{nicefrac}       
\usepackage{microtype}      
\usepackage{xcolor}         
\usepackage{colortbl}

\usepackage{xspace}
\usepackage{comment}
\usepackage{multirow}
\usepackage{adjustbox}
\usepackage{subcaption}  

\usepackage{listings}
\usepackage{color}
\usepackage[most]{tcolorbox}
\usepackage{cleveref}

\definecolor{mygreen}{rgb}{0,0.6,0}
\definecolor{mygray}{rgb}{0.5,0.5,0.5}
\definecolor{mymauve}{rgb}{0.58,0,0.82}

\definecolor{red}{RGB}{255, 102, 102} 

\newcommand{\dataset}{\textsc{CulturalVQA}\xspace}

\newcommand{\nquestions}{2,378\xspace}
\newcommand{\nanswers}{7,206\xspace}
\newcommand{\nunqimages}{2,328\xspace}

\newcommand{\ncountries}{11\xspace}
\newcommand{\avgqlen}{10.98\xspace}
\newcommand{\avgalen}{1.73\xspace}

\crefname{section}{\S}{\S\S}
\crefname{appendix}{App.}{Apps.}
\crefname{figure}{Fig.}{Figs.}
\crefname{table}{Tab.}{Tabs.}

\title{Benchmarking Vision Language Models for Cultural Understanding}

\author{
 \textbf{Shravan Nayak\textsuperscript{1,2}}\hspace{1.2em}
 \textbf{Kanishk Jain\textsuperscript{1,2}}\hspace{1.2em}
 \textbf{Rabiul Awal\textsuperscript{1,2}}
\\[0.5em]
 \textbf{Siva Reddy\textsuperscript{1,3}}\hspace{1.2em}
 \textbf{Sjoerd van Steenkiste\textsuperscript{4}}\hspace{1.2em}
 \textbf{Lisa Anne Hendricks\textsuperscript{5}}
\\[0.5em]
 \textbf{Karolina Sta\'nczak\textsuperscript{1,3}}\hspace{1.2em}
 \textbf{Aishwarya Agrawal\textsuperscript{1,2}}
\\[0.5em]
 \textsuperscript{1}Mila -- Quebec AI Institute,\hspace{0.1em}
 \textsuperscript{2}Université de Montréal,\hspace{0.1em}
 \textsuperscript{3}McGill University,\hspace{0.1em}\\\vspace{0.2em}
 \textsuperscript{4}Google Research,\hspace{0.1em}
 \textsuperscript{5}Google DeepMind
\\[0.3em]
 \small{
   \textbf{Correspondence:} { \tt\href{mailto:shravan.nayak@mila.quebec}{shravan.nayak@mila.quebec}}
 }
}

\begin{document}

\maketitle

\begin{abstract}
Foundation models and vision-language pre-training have notably advanced Vision Language Models (VLMs), enabling multimodal processing of visual and linguistic data. However, their performance has been typically assessed on general scene understanding -- recognizing objects, attributes, and actions -- rather than cultural comprehension.
This study introduces \dataset, a visual question-answering benchmark aimed at assessing VLM's geo-diverse cultural understanding. We curate a collection of \nquestions image - question pairs with 1-5 answers per question representing cultures from \ncountries countries across 5 continents.
The questions probe understanding of various facets of culture such as clothing, food, drinks, rituals, and traditions. Benchmarking VLMs on \dataset, including GPT-4o and Gemini, reveals disparity in their level of cultural understanding across regions, with strong cultural understanding capabilities for North America while significantly lower performance for Africa. We observe disparity in their performance across cultural facets too, with clothing, rituals, and traditions seeing higher performances than food and drink. These disparities help us identify areas where VLMs lack cultural understanding and demonstrate the potential of \dataset as a comprehensive evaluation set for gauging VLM progress in understanding diverse cultures.

\hspace{1.75em}\raisebox{-0.25\height}{\includegraphics[width=1.25em,height=1.25em]{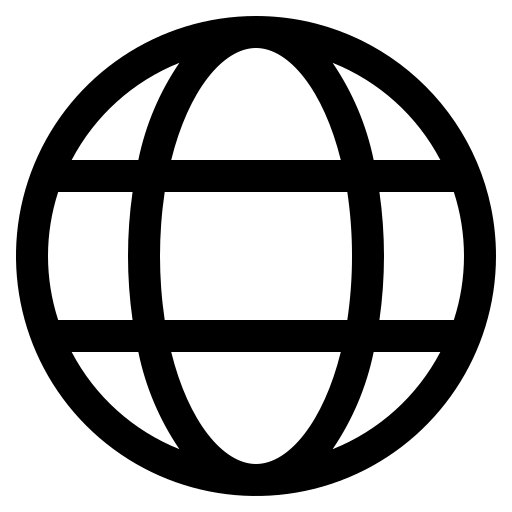}}\hspace{.75em}\parbox{\dimexpr\linewidth-2\fboxsep-2\fboxrule}{\url{https://culturalvqa.org}}
\end{abstract}

\section{Introduction}

Recent multimodal vision-language models (VLMs) \citep{radford21a,liu2023llava,kosmos-2,chen2024far,yujie2024wildvisionarena} have shown impressive performance in tasks such as image-to-text generation, visual question answering, and image captioning, \textit{inter alia}. These tasks predominantly focus on general scene understanding capabilities such as recognizing attributes, objects, and actions in scenes containing objects in their common context \citep{mscoco}. However, given the advancing capabilities of VLMs, we believe the time is now ripe to hold VLMs to higher standards. Indeed, to support increasingly \emph{global} digital interactions, VLMs must also be capable of understanding the \emph{cultural values} \citep{liu-etal-2021-visually} such as beliefs, rituals, and traditions, for a \emph{variety} of cultures in the world.\looseness=-1

\begin{figure}
    \centering
    \includegraphics[width=\linewidth]{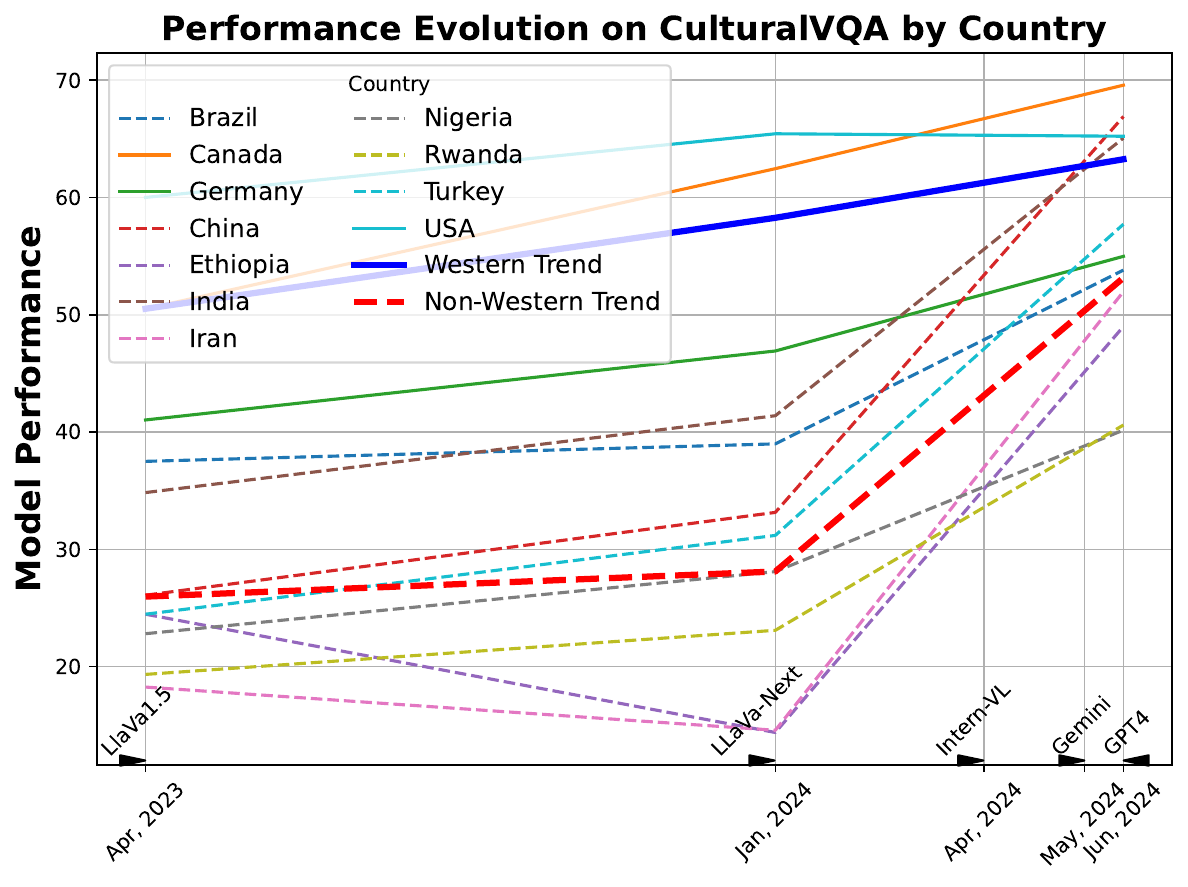}
    \caption{The performance of VLMs over time, segmented by non-Western ({\color{red}red}) and Western ({\color{blue}blue}) countries, with model release dates annotated (bottom). Dashed and solid lines differentiate trends for non-Western and Western countries respectively. VLMs' understanding of non-Western cultures has been in a steep upward trend since Jan '24.}
    \label{fig:teaser}
    \vspace{-10pt}
\end{figure}

\begin{figure*}[ht!]
    \centering
    \includegraphics[width=\linewidth]{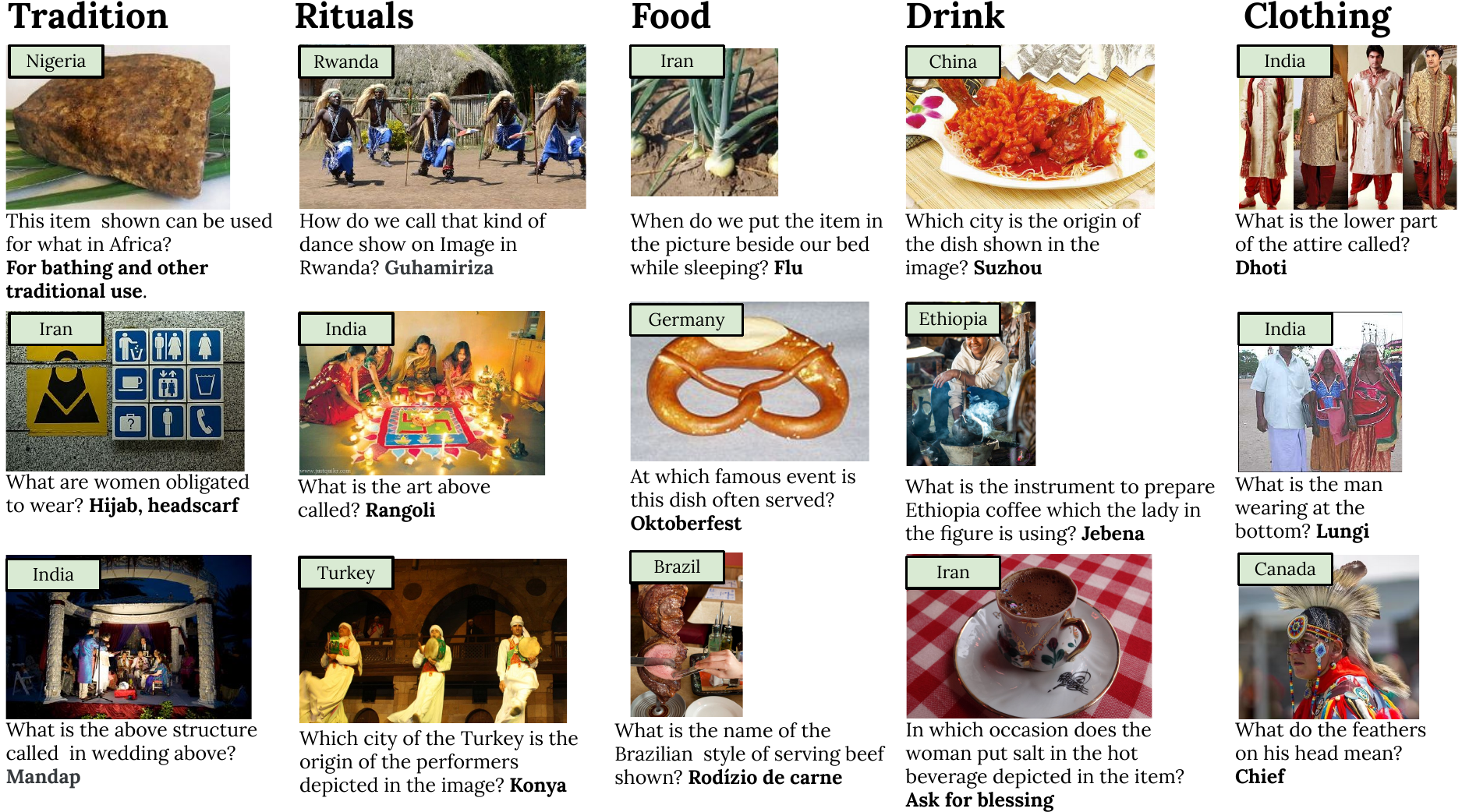}
    \caption{Samples from \dataset. Our dataset is comprised of images presenting cultural concepts from \ncountries countries across five facets: traditions, rituals, food, drink, and clothing. It further includes questions probing cultural understanding of the concepts presented in the images and answers to these questions.}
    \label{fig:vqa_samples}
    \vspace{-10pt}
\end{figure*}

In order to adequately assess whether the current state-of-the-art VLMs -- including proprietary models such as \textsc{GPT-4o} \citep{openai2023gpt4v} and \textsc{Gemini} \citep{gemini} -- encode cultural knowledge, we need systematic benchmarks. However, evaluating cultural understanding is a challenging task since culture is a multifaceted concept consisting of both tangible (e.g., clothing, and food) as well as intangible elements (e.g., ritual practices). Current benchmarks in this domain, including MaRVL \citep{liu-etal-2021-visually} and GD-VCR \citep{yin2021broaden}, while offering foundational insights, have critical shortcomings. 
MaRVL primarily focuses on visual reasoning tasks (e.g., counting, spatial reasoning) on top of images sourced from various cultures, and lacks probing cultural common sense -- the knowledge base
shared by the members of a cultural group (see \Cref{sec:dataset}). While GD-VCR does consider commonsense to a degree, it primarily considers movie scenes, which do not encompass the broader spectrum of everyday cultural contexts.

In response to the above challenges, we propose \dataset, a novel benchmark specifically designed to assess cultural understanding of VLMs.
\dataset is based on Visual Question Answering (VQA), requiring models to integrate both visual and textual information, which permits the formulation of diverse questions, thereby enabling the evaluation of a model's understanding of cultural nuances.
The \dataset benchmark extends the language-only CANDLE dataset \citep{candle2023}, which provides a comprehensive collection of cultural commonsense knowledge assertions. We expand this dataset by automatically collecting images that depict the cultural concept described by the assertions. 
On top of these images, we collect questions and answers by employing annotators from different cultures who would be familiar with the different cultural concepts depicted in the images. See \Cref{fig:vqa_samples} for some examples of questions and answers. Our benchmark consists of \nquestions questions collected on top of \nunqimages unique images with 1-5 answers per question (total \nanswers answers) from \ncountries countries.\footnote{We provide
a data statement in \Cref{app:data-statement}.}
We also present several analyses to better understand the nature of questions and answers in our benchmark.

\begin{table*}[ht]
    \centering
    \begin{adjustbox}{max width=\linewidth}
    \begin{tabular}{lccccccc}
        \toprule
        \textbf{Dataset} & \textbf{No. Regions} & \textbf{No. Questions} & \textbf{No. Images} & \textbf{Multilingual?} & \textbf{Task Format} & \textbf{Culturally Diverse Images?} & \textbf{Nature of Questions} \\
        \midrule
        MaXM \citep{changpinyo2023maxm} & 7 & 2142 & 335 & Yes & Open-ended & No \citep{pouget2024filter} & General reasoning \\
        GDVCR \citep{yin2021broaden} & 4 & 886 & 328 & No & Multiple choice & Yes (movie scenes only) & Cultural understanding \\
        MaRVL \citep{liu-etal-2021-visually} & 5 & 5670 & 4914 & Yes & True/False & Yes & Cultural reasoning \\
        CVQA \citep{romero2024cvqa} & 28 & 9044 & 4560 & Yes & Multiple choice & Yes & Cultural understanding \\
        \dataset (Ours) & 11 & 2378 & 2328 & No & Open-ended & Yes & Cultural understanding \\
        \bottomrule
    \end{tabular}
    \end{adjustbox}
    \caption{Comparison of various datasets closely related to \dataset across different axes.} 
    \label{tab:dataset_overview}
    \vspace{-10pt}
\end{table*}

Further, we systematically evaluate several state-of-the-art VLMs on \dataset. Our evaluation reveals a distinct performance gap between proprietary and open-source models, with open-source models significantly underperforming in comparison (e.g., there is a 29.78\% gap between the highest-performing closed-source model and its open-source counterpart in the country for which the models perform the worst, Ethiopia). Additionally, we observe a significant disparity in model performance across countries. For instance, the highest-performing proprietary model, \textsc{GPT-4o}, achieves 67\% and 72\% accuracy on North American cultural concepts while only between 43\% and 56\% accuracy on concepts from Africa. 
VLMs also show varying degrees of proficiency across cultural facets, with closed-source VLMs performing better on questions about rituals and traditions while scoring worse on those related to clothing, food, and drink.
We develop \dataset as a comprehensive evaluation set for gauging VLMs' progress in understanding diverse cultures and highlighting areas where VLMs lack cultural understanding. We hope that our benchmark will contribute to accelerating the advancements of VLMs in their cultural understanding, as illustrated in \Cref{fig:teaser}.\looseness=-1

\section{Related work}

Cultural understanding is closely related to geo-diverse understanding. For instance, the Dollar Street dataset \citep{dollarstreet} includes 38,479 images of everyday household items from homes around the world, while the GLDv2 dataset \citep{gld} contains 5 million images and 200k distinct instance labels of natural and human-made landmarks, but both only test recognition capabilities as opposed to cultural understanding.
\citet{burdalassen2024culturallyawarevisionlanguagemodels} introduce \textsc{Mosaic}-1.5k, a culture-specific captioning dataset that includes images from various regions. \citet{bhatia2024localconceptsuniversalsevaluating} propose \textsc{GLobalRG}, which aims to evaluate retrieval and grounding capabilities in VLMs across 15 and 50 countries respectively.
Another related line of work focuses on multilingual understanding. For instance, \citet{bugliarello-etal-2022-iglue} unify five datasets across a number of tasks in 20 languages. However, their focus lies in multilingual understanding as opposed to cultural understanding.
Additionally, the XM3600 dataset \citep{ThapliyalCrossmodal2022}, includes image captions from 36 regions and languages, but lacks depth in cultural concepts, making it insufficient for evaluating cultural diversity in VLMs \citep{pouget2024filter}.

Closest to our work are the following benchmarks: MaXM \citep{changpinyo2023maxm}, GD-VCR \citep{yin2021broaden}, MaRVL \citep{liu-etal-2021-visually} and the concurrent work CVQA \citep{romero2024cvqa}. MaXM lacks depth in cultural concepts, as it builds on XM3600 images. Also, its questions focus more on reasoning and general image understanding rather than cultural understanding~\footnote{We manually annotated 100 random questions from the English subset of the MaXM and found the following distribution: color - 3.7\%, spatial understanding - 12.9\%, scene understanding - 42.6\%, Yes/No - 20.4\%, counting - 20.4\% }. The GD-VCR dataset probes cultural understanding, but its reliance on cinematic scenes limits the diversity of real-world cultural contexts it can have. Moreover, they rely on a multiple-choice evaluation format, which can be influenced by the difficulty of answer choices. We believe an open-ended evaluation provides a more faithful assessment of the models' underlying capabilities. Similarly, while MaRVL tests visually grounded reasoning across multiple languages and cultures, it does not assess cultural common sense related to rituals and traditions and also employs a True/False evaluation style. CVQA studies cultural questions in a multilingual setup. However, their focus diverges from ours as they allocate a much smaller proportion of their dataset to traditions and rituals (13\% as compared to 44.1 \% in \dataset) and use a multiple-choice evalaution format. A comprehensive comparison of different datasets across various dimensions is presented in \Cref{tab:dataset_overview}. \dataset uniquely emphasizes open-ended evaluation, includes culturally diverse images (i.e., images from multiple cultures), and  its questions probe for cultural understanding by design. The combination of these characteristics sets \dataset apart from other datasets that either lack culturally diverse images (MaXM), or use restricted evaluation formats such as multiple-choice (CVQA, GDVCR) or True/False (MaRVL).\looseness=-1

\section{\dataset: Dataset Creation}
\label{sec:dataset}
\paragraph{Cultural Taxonomy}
\label{sec:dataset-taxonomy}

Culture is a multifaceted concept that describes the way of life of a collective group of people, distinguishing them from other groups with different cultures \citep{hofstede2010cultures,hershcovich-etal-2022-challenges}. In this paper, we use the concept of a country as a proxy for a cultural group \citep{adilazuarda2024measuring}\footnote{See \Cref{sec:limitations} for a discussion of these choices.}.
Our work assumes common ground within a cultural group by probing \emph{culturally relevant concepts} that are collectively understood, as well as shared \emph{cultural common sense} employed in reasoning \citep{hershcovich-etal-2022-challenges}. For instance, \textit{lavash} -- a traditional Persian bread (see \Cref{fig:vqa_samples}) -- is an example of a culturally relevant concept, while the common practice of \emph{waltzing} at weddings exemplifies the cultural common sense among Germans. 
  
Building on these definitions, we introduce a benchmark that evaluates both the tangible aspects of culture through culturally relevant concepts, such as food, drink, and clothing, as well as the intangible facets via shared common sense embedded in rituals and traditions.\footnote{Herein, the term ‘concepts’ is used to encompass both cultural concepts and common sense.}
We frame this evaluation as a VQA task assessing models' cultural understanding. Starting with a pool of countries, we collect images and use culturally knowledgeable annotators to frame questions. Finally, we collect the ground truth answers.

\paragraph{Selection of Countries}
\label{sec:data-countries}
To build a benchmark that reflects cultural diversity, we aimed to achieve broad geographical coverage. 
Our final dataset spans \ncountries countries and 5 continents. These countries were specifically selected to cover different cultural categories from the World Values Survey \citep{haerpfer2022world} and include Confucian (China), African-Islamic (Turkey, Iran, Ethiopia, Nigeria, Rwanda), Protestant Europe (Germany), English-speaking (USA, Canada), Latin America (Brazil), and South Asian (India) cultures. We opt for an intentional overrepresentation of African-Islamic countries to address their typical scarcity in geo-diverse datasets.

\paragraph{Selection of Images}
\label{sec:data-images}
We use the CANDLE dataset \citep{candle2023} for our image source which contains 1.1 million entries of Cultural Commonsense Knowledge (CCSK) along with URLs to corresponding  webpages from the C4 corpus \citep{10.5555/3455716.3455856}. The CANDLE dataset represents cultural concepts from approximately 196 countries and 80\% of web pages in this corpus contain images related to the text \citep{mmc4}. This allows us to begin with a culturally relevant pool of images.

We apply filters for aspect ratio, size, and specific keywords to refine the image dataset. Further, we use CLIP similarity \citep{hessel-etal-2021-clipscore} to filter images for cultural relevance, discarding those with a CLIP score below a threshold  determined through qualitative evaluation of sample images.\footnote{Threshold of 23 (precision = 0.92, recall = 0.96)} Since our initial pool already contains culturally relevant images, there is minimal risk of introducing western-centric biases through the use of CLIP, despite potential biases in its pretraining data. To further ensure quality, we apply an additional round of human filtering (detailed in the next section). Thus, our multi-stage filtering ensures that the final set of images is appropriate for cultural annotations. Further details of the image filtering process are provided in \Cref{ref:app-instructions-image}.

\begin{figure*}[ht!]
    \centering
    \includegraphics[width=\linewidth]{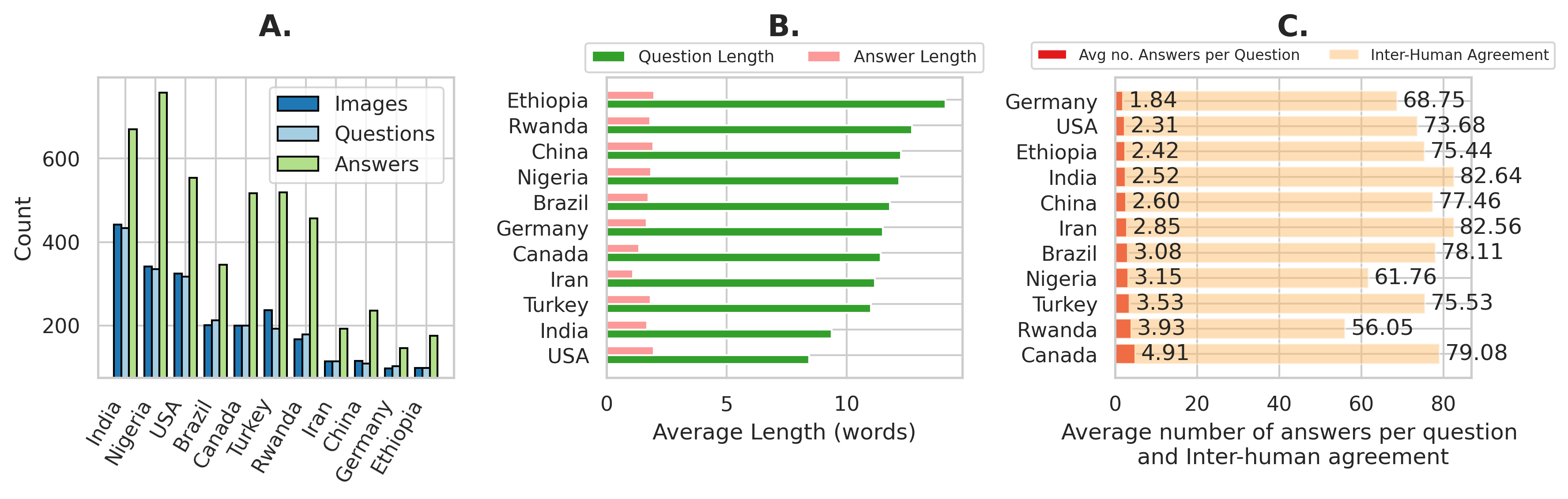}
    \caption{Comparative analysis of data by country. The figure presents three aspects: (A) unique counts of images, questions, and answers, (B) average lengths of questions and answers, and (C) average number of answers per question and inter-annotator agreement scores across countries, showcasing variations and trends in \dataset.}
    \label{fig:unique_counts}
    \vspace{-5pt}
\end{figure*}

\begin{figure*}[ht!]
\includegraphics[width=\linewidth]{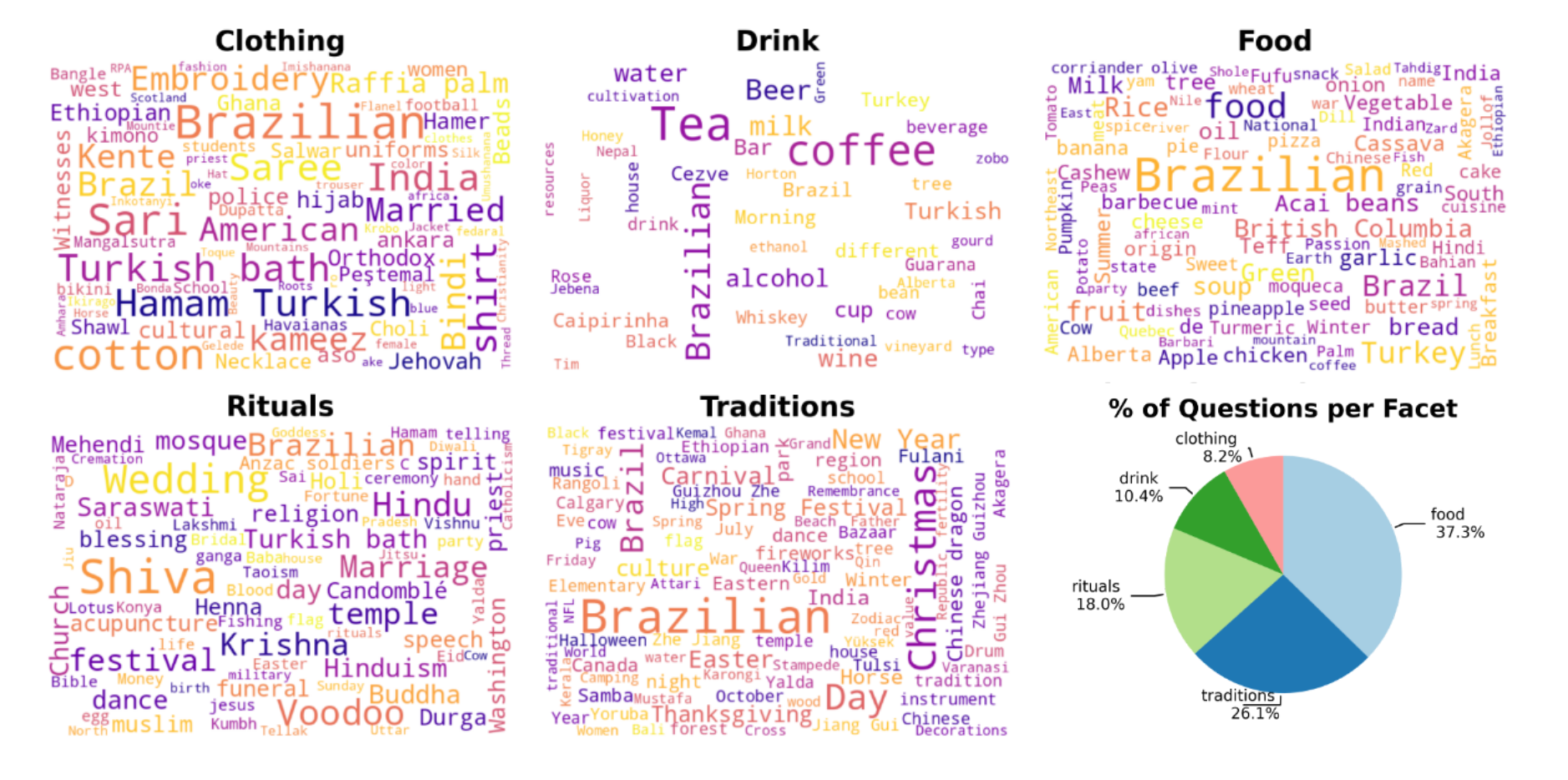}
\caption{Word clouds representing the answers in \dataset across five facets of culture: clothing, drink, food, rituals, and traditions. In the bottom right, a breakdown of cultural facets in data is depicted.}
\label{fig:qa_facet_distribution}
\vspace{-5pt}
\end{figure*}

\paragraph{Question Collection}
\label{sec:data-q}

Following the conceptual culture framework by \citet{hofstede2010cultures},  
we direct annotators to create questions that are easily answerable by someone from their own culture but challenging for outsiders. 
To elicit such questions, we provide annotators the instructions shown in \Cref{app:instructions-quest} as well as images and additional context for the cultural concepts present in the image (retrieved from CANDLE). We encourage them to create questions based on their own cultural knowledge, using the additional context (accessible behind a click-to-expand box) only when absolutely necessary. We also advise annotators to skip images if they found them culturally irrelevant or were unfamiliar with the depicted content. This adds an additional layer of filtering, resulting in annotators discarding 19.64\% of the images shown to them.\looseness=-1 

\paragraph{Answer Collection}
\label{sec:data-ans}

Next, we ask the annotators to write answers to the questions created in the previous step, while ensuring that the answers reflected common agreement within their culture (see instructions in \Cref{app:instructions-answer}).
Here we prompt them to use English for universal concepts like \textit{cats} or \textit{apples} and use widely recognized and agreed upon local terms for concepts like festivals or local cuisine, rather than translating them into English. For example, the annotators should write the term \textit{Naan} instead of \textit{Indian bread}.
This approach preserves the cultural specificity of the collected answers. 
Further, we instruct annotators to be as precise as possible in their answers (e.g., \textit{sushi} instead of \textit{food} and \textit{Oolong tea} instead of \textit{tea}) and to keep their responses concise, ideally between one to three words.\looseness=-1 

Further details about the rationale behind the data curation process and the challenges encountered are provided in \Cref{app:challenges}.\looseness=-1 

\section{Dataset Analysis}
\label{sec:dataset-analysis}

This section provides a detailed analysis of {\dataset}s' composition and characteristics including analysis of images, questions, answers, and cultural concepts contained in it.

\paragraph{Images} \dataset comprises \nunqimages unique images. In \Cref{fig:vqa_samples} we show representative samples.
We choose images to ensure significant cultural representation across \ncountries different countries. The distribution of unique image count per country is detailed in \Cref{fig:unique_counts} (A).

\paragraph{Questions} We collect \nquestions questions in total. In \Cref{fig:unique_counts} (A), we present the number of unique questions per country. The questions have an average length of \avgqlen words (see \Cref{fig:unique_counts} (B) for country-wise breakdown).
Most frequent question types include `What'(51.3\%), `Which'(11.2\%), `In' (5.6\%), and `Why' (3.4\%) questions. 
For example, `What' questions often relate to identifying cultural entities like \textit{saree} or \textit{Dirndl} (traditional Indian and German dresses, respectively) in the clothing category, or festivals like \textit{Spring Festival} (celebrated in China) among rituals. `Where' questions inquire about locations significant to specific foods, such as the origins of \textit{Quebec chicken}. 
Finally, we analyze whether the collected questions contain stereotypes and found that they are largely absent (see \Cref{app:stereotypes}).\looseness=-1 


\paragraph{Answers}\label{sec:answers} \dataset consists of \nanswers manually curated answers in total.\footnote{We collect 1-5 answers per question, depending on the availability of annotators.} The average answer length is \avgalen words (see \Cref{fig:unique_counts} (B) for country wise breakdown). We assess whether answers predominantly feature terms from local languages. To this end, we verified how many answers have corresponding English Wikipedia titles;  
for 80\% of the answers at least one of the answer words is contained in at least one Wikipedia title. Thus our benchmark is still suitable for English VLMs.

\paragraph{Cultural Concepts}  
According to the pie chart in \Cref{fig:qa_facet_distribution}, food-related questions are most prevalent, accounting for 37.3\% of the dataset, followed closely by traditions and rituals, which represent 26.1\% and 18\% respectively. Thus, roughly 44\% of the questions in our dataset probe for cultural understanding of the intangible aspects of culture (rituals and traditions). The \textbf{word clouds} generated from the collected answers in \Cref{fig:qa_facet_distribution} illustrate the diversity of expressions, such as hamam (Turkey) and meskel (Ethiopia) for rituals and traditions, and feijoada (Brazil), fufu (Nigeria), and vada (India) for food, indicating a geographically diverse culinary and cultural scope. While the clothing category is the least prevalent in the dataset, it shows the highest variety in terms of collected answers.
The drink category is notably one of the smallest, both in terms of the size and number of unique answers.

\section{Evaluating VLMs on \dataset}

\paragraph{Evaluation Metric}

Evaluating open-ended VQA is challenging. Traditionally, string matching has been used but it is known to underestimate model performance. Based on findings from \citet{manas2024improving}, which demonstrate the effectiveness of reference-based LLM evaluation for open-ended VQA tasks, we adopt LAVE, their proposed metric, as our evaluation metric with \textsc{GPT-4} as the LLM (see \Cref{app:lave} for the LLM prompt used). We validated the effectiveness of LAVE for our use case by computing correlation with human judgments. LAVE judgment agrees with human judgment 79\% of the times for \textsc{GPT-4}, 73\% of the times for \textsc{Gemini}, and 76\% of the times for \textsc{Intern-VL}.

\paragraph{VLMs used for evaluation}

We benchmark several state-of-the-art VLMs on the proposed \dataset dataset, ranging from closed-source models like \textsc{GPT-4} (\textsc{GPT-4o}), \textsc{Claude} (\textsc{Claude 3.5}) and \textsc{Gemini Pro} (\textsc{Gemini-Pro-Vision 1.0}) to a wide variety of open-source models, ranging from 7 to 25 billion parameter count:
\textsc{Blip2}~\cite{li2023blip},
\textsc{InstructBlip}~\cite{dai2024instructblip},
\textsc{mBLIP}~\cite{geigle2023mblip}
\textsc{PalliGemma}~\cite{beyer2024paligemmaversatile3bvlm}
\textsc{Llava1.5}~\cite{liu2023llava}, \textsc{Llava\_Next}~\cite{liu2024llavanext}, \textsc{Idefics2}~\cite{laurenccon2024matters}, and \textsc{Intern-VL 1.5}~\cite{chen2024far}. See \Cref{app:models} for a detailed discussion of the selected models.\looseness=-1

\begin{table*}[ht]
\small
\centering
\begin{adjustbox}{max width=\linewidth}
\begin{tabular}{@{}l@{\hskip 0.15cm}ccccccccc@{}}
\toprule
         & \multicolumn{6}{c}{\textbf{Open-Source}}                         & \multicolumn{3}{c}{\textbf{Closed-Source}} \\ \cmidrule(lr){2-7} \cmidrule(lr){8-10}
\bf Country & \textsc{mBLIP} & \textsc{LLaVa1.5} & \textsc{Blip2} & \textsc{LLaVa-Next} & \textsc{Idefics2} & \textsc{Intern-VL} & \textsc{Gemini} & \textsc{Claude} & \textsc{GPT-4} \\\midrule

\textbf{Brazil  } & 25.34 & 40.38 & 32.21 & 45.62 & \cellcolor{blue!25}54.37 & 52.53 & 66.34 & 66.36 & \cellcolor{green!25}76.44 \\
\textbf{Canada  } & 38.50 & 50.50 & 58.50 & 62.50 & \cellcolor{blue!25}69.00 & 67.50 & 65.50 & 66.00 & \cellcolor{green!25}72.00 \\
\textbf{China   } & 22.61 & 26.09 & 34.78 & 33.04 & 38.26 & \cellcolor{blue!25}53.04 & 65.22 & 49.57 &\cellcolor{green!25} 65.22 \\
\textbf{Ethiopia} & 7.44 & 24.47 & 17.02 & 18.09 & 25.53 & \cellcolor{blue!25}26.60 & 42.55 & 41.49 &\cellcolor{green!25} 56.38 \\
\textbf{Germany } & 41.02 & 41.03 & \cellcolor{blue!25}51.28 & 48.72 & 38.46 & 48.72 & 48.72& 51.28 & \cellcolor{green!25}61.54 \\
\textbf{India   } & 27.83& 34.84 & 46.61 & 42.53 & 49.32 & \cellcolor{blue!25}53.85 & 58.37 & 59.28 &\cellcolor{green!25} 69.68 \\
\textbf{Iran    } & 13.04 & 18.26 & 19.13 & 17.39 & 23.48 & \cellcolor{blue!25}30.43 & 46.09 & 47.83 &\cellcolor{green!25} 57.39 \\
\textbf{Nigeria } & 13.16 & 22.81 & 21.35 & 28.95 & 31.87 & \cellcolor{blue!25}33.92 & 36.26 & 36.55 &\cellcolor{green!25} 43.27 \\
\textbf{Rwanda  } & 13.26 & 19.34 & 22.65 & 25.41 & 23.20 & \cellcolor{blue!25}28.73 & 35.36 & 33.70 &\cellcolor{green!25} 46.41 \\
\textbf{Turkey  } & 28.57 & 24.47 & 33.76 & 33.33 & 37.97 & \cellcolor{blue!25}41.35 & 56.12 & 51.26 & \cellcolor{green!25}59.92 \\
\textbf{USA}      & 38.77 & 58.77 & 62.77  & 62.77 & 65.23 & \cellcolor{green!25}71.38 & 62.15 & 64.92 &  66.77   \\ \midrule
\textbf{Average} & \textit{24.50} & \textit{32.81} & \textit{36.37} & \textit{38.03} & \textit{41.51} & \textit{46.18} & \textit{52.97} & \textit{51.66} & \textit{61.36} \\
\bottomrule
\end{tabular}
\end{adjustbox}
\caption{LAVE accuracies of open- and closed-source models on \dataset. Best-performing results per country are highlighted in {\color{green!25} green}, and best-performing results among open-source models are highlighted in {\color{blue!25} blue}.}

\label{tab:country_performance}
\vspace{-10pt}

\end{table*}

\begin{figure}[ht!]
    \centering
    \includegraphics[width=\linewidth]{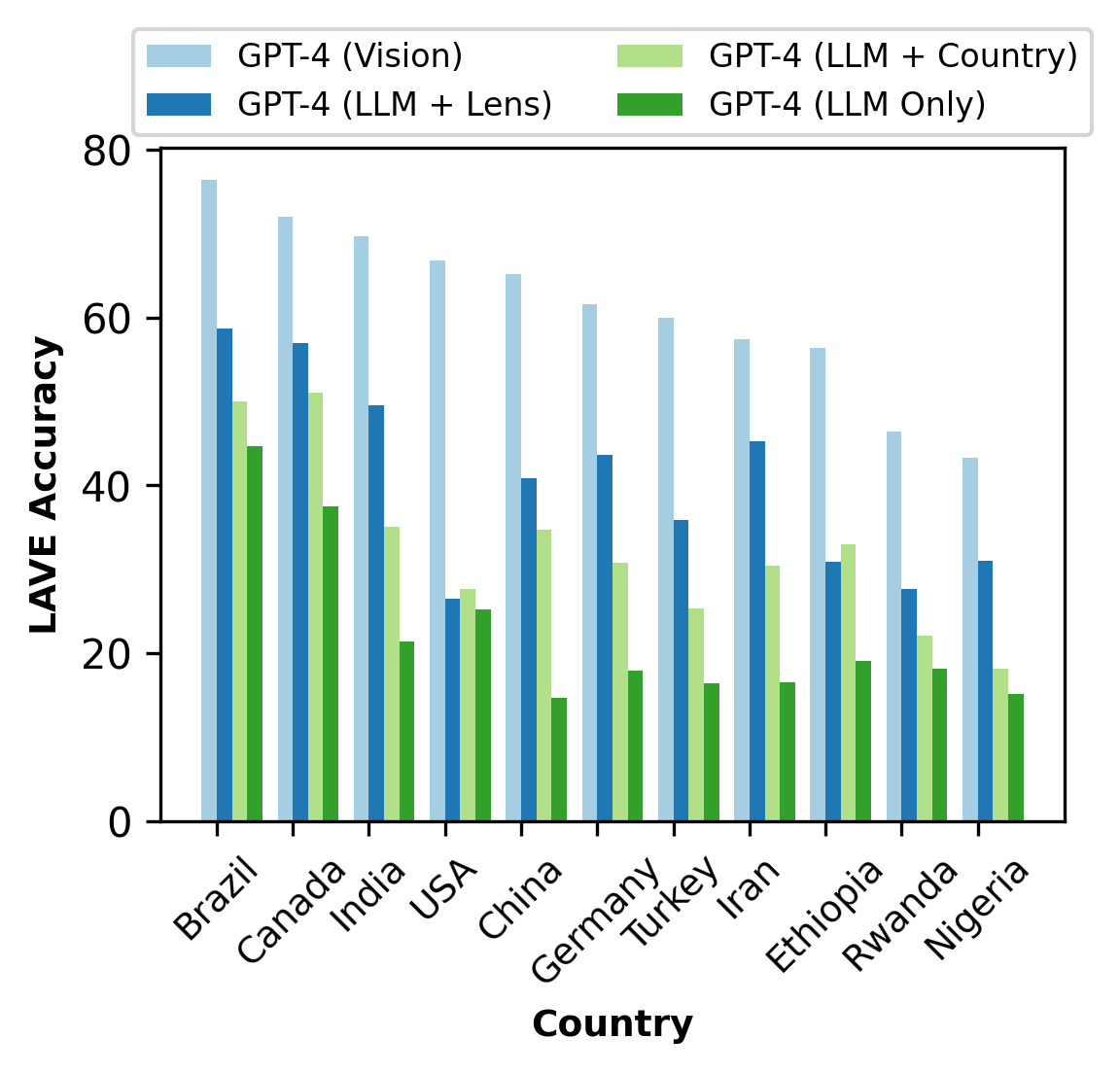}
    \caption{Baseline evaluation of the degree of visual understanding required in \dataset: LLM-only, LLM with a country-specific context, LLM with Google Lens entities, and \textsc{GPT-4V}.}
    \label{fig:model-perf}
    \vspace{-15pt}
\end{figure}

\paragraph{What degree of visual understanding is required to answer the questions in \dataset?} To investigate this, we employ the following baselines. 
\textbf{LLM-only:} This baseline uses an LLM to answer questions based solely on the question input. It helps gauge how well the questions can be addressed without visual context, relying only on the cultural knowledge encoded in the LLM. 
\textbf{LLM + Country:} It introduces country-specific context into the LLM prompts to determine if knowing the country along with the question can already elicit the correct answer.
\textbf{LLM + Lens:} This baseline uses image entity names extracted by Google Lens \citep{googlelensapi} along with the question as input, unlike the other baselines that lack visual context. It helps assess whether coarse-level visual knowledge is sufficient to answer the questions. 

We evaluate the baselines using \textsc{GPT-4} as the underlying LLM. The LAVE accuracies for these baselines, as well as for the \textsc{GPT-4} VLM (which also incorporates an image as input in addition to the question), are presented in \Cref{fig:model-perf}. We see that although the country information and the coarse visual entities help improve the performance on top of the LLM-only baseline, the performance of the strongest baseline (LLM + Lens) is still far from that of the VLM.
This verifies that the questions in our dataset require sufficient visual understanding to answer them accurately.

\begin{figure}[ht!]
    \centering
    \includegraphics[width=\linewidth]{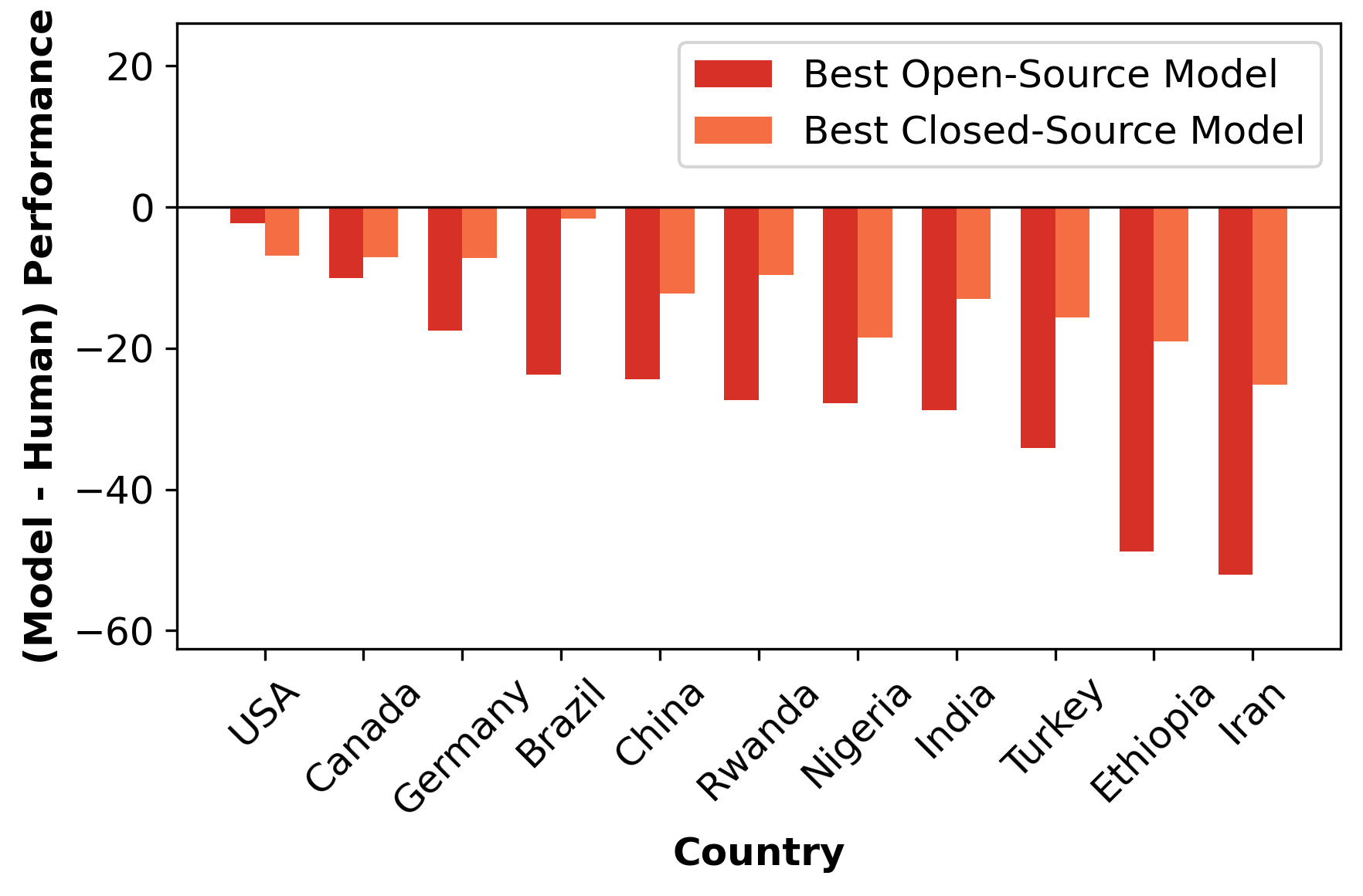}
    \caption{Performance gap between the best open-source (one of \textsc{InternVL}, \textsc{Idefics2}, \textsc{BLIP2}) and closed-source models (\textsc{GPT-4o}) compared to human performance. Negative values indicate where models underperform relative to humans.}
    \label{fig:comp_gpt4_human}
    \vspace{-15pt}
\end{figure}

\paragraph{To what extent are VLMs culturally aware?} 


We report the LAVE accuracies for \textbf{zero-shot} evaluation of VLMs on the proposed \dataset benchmark in \Cref{tab:country_performance} and \Cref{tab:model_performance_extra}. The average LAVE accuracy for the highest-performing model, \textsc{GPT-4}, is approximately 61\%, with performance varying across countries from 43\% to 72\%. We see substantial disparity in cultural understanding across different VLMs, with the best-performing open-source model (\textsc{Intern-VL} for most countries) achieving an average LAVE accuracy of only 46\%, and performance ranging across countries from 26\% to 71\%.
This result indicates a considerable performance gap between closed-source models and the best-performing open-source model. It is particularly pronounced in countries within the African-Islamic culture (Ethiopia, Nigeria, Iran, and Turkey), with a 29.78\% gap for Ethiopia, the country for which the models perform the worst. We also conduct \textbf{few-shot} evaluation of VLMs but find that it does not significantly impact performance (see \Cref{app:few_shot} for more details). Hence, the subsequent analyses in this section are conducted on top of zero-shot results.\looseness=-1 

\begin{figure}[]
    \centering
    \includegraphics[width=\linewidth]{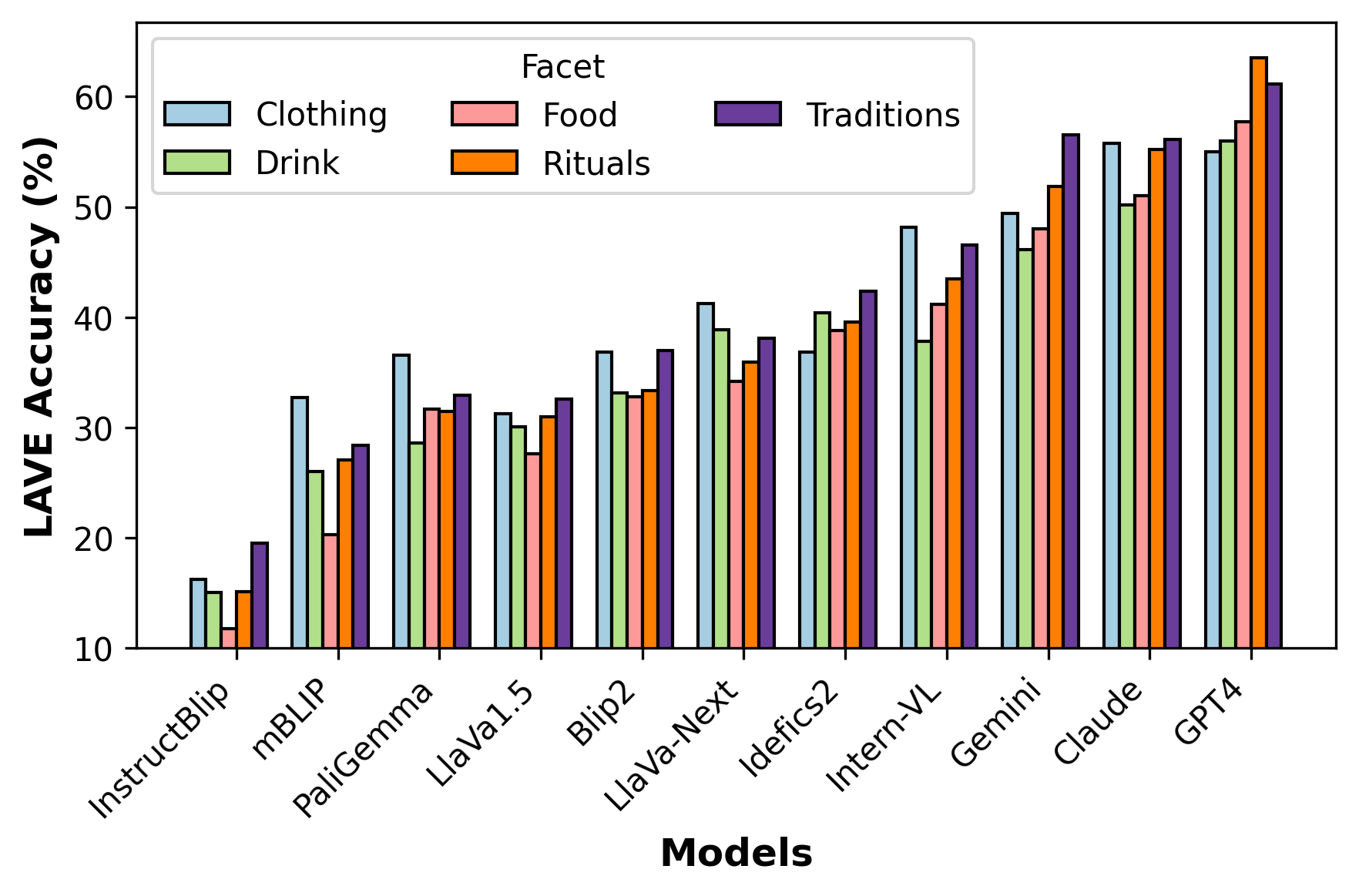}
    \caption{VLM performance across facets as measured using LAVE accuracies.}
    \label{fig:facet}
    \vspace{-15pt}
\end{figure}

\paragraph{Are VLMs better at understanding cultures from some countries than others?} A country-level (see \Cref{tab:country_performance}) analysis of the models reveals stark variance in performance across different regions. Generally, open-source models perform well for high-resource countries such as the USA, Canada, Brazil, and India while achieving inferior performance in underrepresented countries (Ethiopia, Iran, and Rwanda). This trend holds true even for open-source models with large parameter sizes, such as \textsc{Intern-VL}, indicating that data diversity is more crucial for cultural understanding than model size. Although closed-source models showcase less drastic performance discrepancies across countries, their performance also degrades significantly for African-Islamic countries.

\paragraph{Are VLMs better at understanding some cultural concepts than others?} In \Cref{fig:facet}, we report the model performance across 5 cultural facets.
Generally, we find that proprietary models tend to perform better on intangible concepts -- rituals, and traditions, compared to drink and food. Indeed, the highest performance of \textsc{GPT-4} is achieved in the rituals facet ($\approx 63\%$), whereas in the clothing facet, it achieves a lower performance of $\approx 55\%$. Refer to \Cref{app:facets} for a more detailed discussion.

\paragraph{Do multilingual VLMs perform better in culturally diverse settings?} One might expect that multilingual VLMs may demonstrate superior performance due to their exposure to culturally diverse data. However, our analysis of multilingual models, mBLIP and PaliGemma, on \dataset reveals a more nuanced picture. From \Cref{tab:country_performance}, mBLIP, built on top of monolingual BLIP2, consistently underperforms it despite multilingual training. This could be due to the quality of the machine-translated data used in mBLIP and the LLM backbone used (mT0 \citep{muennighoff2023crosslingualgeneralizationmultitaskfinetuning} in mBLIP vs. FlanT5 \citep{chung2022scalinginstructionfinetunedlanguagemodels} in BLIP2).
Also, from \Cref{tab:model_performance_extra} we observe that PaliGemma shows significant disparities across countries despite large-scale multilingual training. This is possibly due to its smaller size (3B) which suggests that multilingual data exposure alone is insufficient for cultural understanding. 

\begin{figure*}[ht!]
\centering
\includegraphics[width=\linewidth]{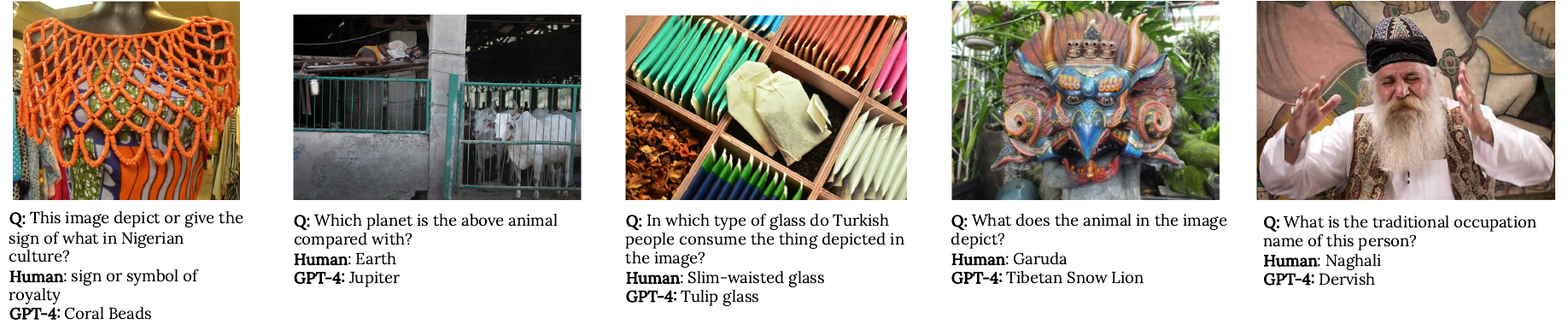}
\caption{Qualitative failure examples of \textsc{GPT-4} predictions.}
\label{fig:errors}
\vspace{-15pt}
\end{figure*}

\paragraph{How do culturally knowledgeable people perform on \dataset?} We calculate human performance for 1,455 questions for which we have three or more answers using the LAVE metric.
For each question, we compute the accuracy of one of the human answers against the remaining human answers using LAVE. We then average the scores across all answers. 
Since all these answers are written by annotators who are familiar with the culture probed in the question, this human performance tells us how well culturally knowledgeable people perform on \dataset.

Based on the results reported in \Cref{fig:unique_counts} (C), human performance is notable and ranges from 55\%-85\%, with certain countries, such as Iran, showing particularly high scores ($>80\%$). In contrast, Rwanda and Nigeria had the lowest scores (56.05\% and 61.76\%, respectively). These lower scores can be partially attributed to the cultural diversity within these countries, where using a country as a proxy for a cultural group may not accurately capture the nuances of subcultural variations. The same concept may hold different meanings across subcultures, leading to varied interpretations and inconsistencies in responses. Further qualitative insights are provided in the \Cref{app:human_performance}.

We also calculate the Pearson correlation between human and model performance across countries. For open-source models, we observe a relatively low correlation, ranging from 0.1 to 0.4. Interestingly, for closed-source models like \textsc{Gemini} and \textsc{GPT-4}, we find a stronger correlation of 0.69 and 0.75, respectively. This suggests that the factors affecting human performance similarly influence the performance of these closed-source models.
However, from \Cref{fig:comp_gpt4_human}, when comparing human and model performance using the same metric, we find that closed-source models still lag behind humans for \textit{all countries} indicating that while these models follow human performance trends, there is still a marked gap in their cultural understanding compared to humans. This gap is even more pronounced for open-source models, which show an even larger discrepancy across \textit{all countries}.

Further, in \Cref{fig:comp_gpt4_human}, we observe a larger gap for non-Western countries such as Iran, Nigeria, India, Turkey, and Ethiopia ($>13\%$). Conversely, the smaller gap for Canada and the USA (< 7.0\%) indicates a closer alignment between models and human performance, likely due to a better representation of Western cultural concepts in the training data. Interestingly, GPT-4 shows a relatively low gap for Brazil ($\approx 2\%$), possibly because the questions for Brazil often probe coarse visual understanding. This trend is further supported by LLM + Lens baseline in Figure \ref{fig:model-perf} which performs exceptionally well for Brazil. \looseness=-1

\paragraph{How much does varying question difficulty and varying answer counts affect model performance disparity across countries?}

Since we sourced questions from different annotator groups across countries, it is imperative to ask if the disparity in model performance across countries is due to differences in inherent question difficulty across countries. To investigate this, we analyze the Spearman rank correlation~\footnote{\label{spearman_reason}We use this metric as the variance in lengths is small, making rank-based analysis more meaningful.} between the model performance and the average question length (see \cref{fig:unique_counts} (B) for average question length across countries). We use average question length as a proxy for question difficulty - assuming shorter questions probe more direct knowledge, while longer ones require nuanced cultural understanding. We found a weak correlation between question length and model performance (-0.3 to 0.3) for most models, with the exception of \textsc{GPT-4} and \textsc{Gemini}, which showed a moderate negative correlation of -0.52 on average. As illustrated in \Cref{fig:corelation}, for most countries except Brazil, Canada, and the USA, the variance in question lengths is small, suggesting that question length is not a significant factor behind the disparity in model performance across countries.\looseness=-1

Another factor potentially affecting the disparity in model performance is the variable number of human answers per question across countries (see \Cref{fig:unique_counts} (C)). These human answers are used as the reference answers in the LAVE metric making it more rigid for countries with fewer references and vice-versa. To investigate this, we compute the Spearman correlation\footref{spearman_reason} between model performance and the average number of answers per question across countries. We find a very low correlation ranging between -0.3 and 0 across models, indicating that the disparity in the number of human answers does not meaningfully affect the disparity in model performance across countries.

\paragraph{Human judgment of model performance} 
We evaluate responses from the three best-performing models, \textsc{GPT-4}, \textsc{Gemini}, and \textsc{Intern-VL} to questions from India, with each answer rated by 5 humans on a scale of 1 to 5, from completely correct to completely incorrect.
See \Cref{app:human-judgment} for details on the human evaluation study.
\Cref{fig:hum_jud} shows the percentage of questions that fall into each of the five scales.
The models' scores closely align with human judgments for case 1 scores, suggesting that our metric predicts answers to be correct only if they are both precise and culturally specific. We note that humans tend to rate model predictions higher than the LAVE metric.
Finally, the evaluation shows that humans tend to choose the extreme ratings,
considering most model responses as either fully accurate or entirely wrong.

\paragraph{Qualitative examples of model failures}
Our qualitative evaluation of the best-performing model, \textsc{GPT-4}, highlights its limitations in recognizing and interpreting cultural nuances. 
For instance, \textsc{GPT-4} overlooks the cultural significance of intangible cultural concepts like coral beads in Nigeria, which symbolize wealth and heritage but are treated merely as decorative objects, as well as it fails to recognize the symbolic connection between cows and planet Earth in Indian culture (see \Cref{fig:errors}).
Focusing on tangible cultural concepts in \Cref{fig:errors}, the model's shortcomings are evident as it inaccurately recognizes cultural entities and objects. For instance, it mislabels 
\textit{Naghali}, a traditional Iranian storyteller as a Dervish and mistakes a traditional Turkish tea glass for a tulip glass, commonly used for serving beer.
These examples reveal how \textsc{GPT-4} 
has difficulties distinguishing between visually similar but culturally distinct entities and objects, and it lacks a deep understanding of cultural beliefs and symbolic meanings.\looseness=-1

\section{Conclusions}

In this paper, we introduce \dataset, a novel VQA benchmark for assessing VLMs on their cultural understanding.
By curating a diverse collection of images from \ncountries countries across 5 continents and collecting \nquestions hand-crafted questions and \nanswers answers about cultural concepts presented in these images, written by annotators, we ensured a broad representation of cultural concepts pertinent to diverse cultural groups.

Benchmarking state-of-the-art models on \dataset reveals notable disparities in their performance across regions. Models perform much better on North American cultures compared to African-Islamic ones. Further, we find a stark performance disparity between closed- and open-source models, with a 29.78\% gap between the highest-performing closed-source and open-source models for the lowest-performing country. VLMs also show varying proficiency across cultural facets, excelling in questions about clothing, rituals, and traditions but struggling with food and drink. Our results underscore the current limitations of VLMs in achieving uniform cultural comprehension and pinpoint specific areas that require improvement.

\section{Limitations}
\label{sec:limitations}
Our study faces limitations due to our data collection methods, the scope of the \dataset dataset, and our focus on the English language. We approximated cultural groups using geographical regions for annotator recruitment, potentially oversimplifying cultural identities and conflating culture with nationality due to practical constraints like annotator availability. We acknowledge that some cultural concepts may lack local terms that can be effectively represented in English letters \footnote{https://tinyurl.com/3zvjsv6x}. Hence, our use of English-only data may also miss key cultural nuances available only in native languages. In such cases, collecting annotations in native languages would help mitigate this issue. However, we emphasize that our benchmark, despite being in English, is already challenging enough for the models, as evidenced by the significant disparity in model performance across cultures. In \Cref{sec:answers}, our analysis revealed that 80\% of the answers contain at least one word matching an English Wikipedia page, while 20\% lack such a match. This suggests that these answers may be multilingual, which presents a limitation for our English-only benchmark. Although our dataset aims for cultural diversity, it does not capture the full spectrum of global cultural diversity. Future work will expand the dataset to represent diverse cultures and regions more broadly and develop multilingual datasets for greater inclusivity.\looseness=-1

\paragraph{\textbf{Challenges in collecting culturally informative data}} Collecting culturally rich content from annotators proved challenging, particularly because the images and concepts were limited to those available on English-language websites. This restriction likely omits important cultural details. Allowing annotators to skip inadequate images did not fully overcome the drawbacks of limited image quality, impacting the depth of the questions created.

\section{Ethical Considerations}

Our \dataset benchmark involves culturally specific questions and answers, developed by professional annotators from the relevant countries. We sought wide cultural representation by engaging with three different communities, compensating annotators at \$10-15 per hour for both included and excluded contributions after pilot testing. This reflects our best effort to maintain fairness and inclusivity in our data collection process.

Despite these efforts, we recognize our approach’s limitation in equating cultural groups with national borders, potentially overlooking the complex realities of minority and diaspora communities. We urge future research to explore finer distinctions within cultural groups to enhance representation. Although we have rigorously tried to remove biases, some subjective content may persist; however, a substantial portion of the dataset has been verified as unbiased (see \Cref{app:stereotypes}). We acknowledge these constraints but are hopeful that our work will advance the understanding of cultural nuances in VLMs.\looseness=-1

\section{Acknowledgements}
We would like to extend our gratitude to David Ifeoluwa Adelani for connecting us with Masakhane, Fırat Öncel for assisting with annotators in Turkey, Saba Ahamadi for securing annotators from Iran, and Qian Yang for sourcing annotators from China. We also appreciate the valuable feedback provided by Ibrahim Alabdulmohsin on the early draft. The technical support from the Mila IDT team in managing the computational infrastructure is greatly appreciated. We would also like to thank Chris Emezue for his assistance in helping with payments for the annotators. Additionally, Aishwarya Agrawal received support from the Canada CIFAR AI Chair award throughout this project. Karolina Sta\'nczak was supported by the Mila P2v5 grant and the Mila-Samsung grant. This project was generously funded by a research grant from Google.

\small
\bibliography{main}

\normalsize
\appendix

\section*{Appendix} 

\addcontentsline{toc}{section}{Appendices}
\renewcommand{\thesubsection}{\Alph{subsection}}

\subsection{Data Statement}
\label{app:data-statement}

We provide a data statement \citep{bender-friedman-2018-data} to document the generation and provenance of \dataset.

\begin{table*}[]
\centering
\small
\begin{adjustbox}{max width=\linewidth}
\begin{tabular}{lccccccccccc}
\toprule
\textbf{Country}        & Brazil & Canada & China & Ethiopia & Germany & India & Iran & Nigeria & Rwanda & Turkey & USA \\ \midrule
\textbf{No. Annotators} & 5      & 6      & 6     & 4        & 11      & 6     & 4    & 8       & 7      & 4      & 5 \\ \bottomrule
\end{tabular}
\end{adjustbox}
\caption{Number of Annotators by Country}
\label{tab:annotators}
\end{table*}

\begin{table*}[!ht]
\centering
\begin{adjustbox}{max width=\linewidth}
\begin{tabular}{lcccccccccccc} \toprule
Model        & Brazil & Canada & China & Ethiopia & Germany & India & Iran  & Nigeria & Rwanda & Turkey & USA   & Avg.   \\ \midrule
PaliGemma    & 38.87  & 54.50  & 20.87 & 9.57     & 35.89   & 35.52 & 13.04 & 19.88   & 14.36  & 26.05  & 54.77 & \textit{28.67} \\
InstructBLIP & 10.57  & 17.00  & 16.52 & 3.19     & 30.77   & 19.91 & 11.30 & 13.74   & 4.97   & 21.52  & 29.54 & \textit{16.27} \\ \bottomrule
\end{tabular}
\end{adjustbox}
\caption{Performance of InstructBLIP and PaliGemma on CulturalVQA}
\label{tab:model_performance_extra}
\end{table*}

\paragraph{Curation Rationale}

\dataset benchmark is designed to evaluate VLMs' cultural understanding capacities across various cultures. 
The images are sourced from the CANDLE dataset \citep{candle2023}, which offers a comprehensive collection of Cultural Commonsense Knowledge (CCSK) from the C4 corpus \citep{10.5555/3455716.3455856}, consisting of 1.1 million entries each linked to relevant CCSK data via URLs to webpages.
Annotators writing questions and answers for this project are recruited through the MTurk platform, an African NLP organization, and an international academic AI research institute. 

\paragraph{Language Variety} 
All texts in the dataset are in English, primarily authored by non-native speakers, and may contain ungrammatical structures in both questions and answers. We build our dataset in English to disentangle multicultural understanding from multilingual comprehension.

\paragraph{Annotator Demographics}
All annotators come from the following \ncountries countries: China, Turkey, Iran, Ethiopia, Nigeria, Rwanda, Germany, USA, Canada, Brazil, and India. Initially, we attempted to engage professional annotators from the Amazon Mechanical Turk (MTurk) platform. However, we encountered challenges in finding sufficient presence of annotators from some of the targeted countries. Therefore, we expanded our search to other communities with a broad cultural representation, including Masakhane, an African NLP organization, and Mila, an international academic AI research institute. All annotators are either natives of the country they annotated for or have resided there for at least 18 years, ensuring they have sufficient cultural context and lived experiences required for the task. We conducted multiple pilot rounds to ensure that annotators adhere to our guidelines and are fluent in English. Other demographics such as age and gender are unknown. All annotators were compensated at an hourly rate of 10-15\$ per hour depending on a task and the number of completed HITs. The number of unique annotators from each country can be found in \Cref{tab:annotators}.

\subsection{Image Filtering}
\label{ref:app-instructions-image}
Given the potential noise inherent in an image dataset derived from web scraping, we implement heuristic filters to refine our selection. First, we apply aspect ratio filtering, retaining only images with an aspect ratio between 0.5 and 2, effectively removing many banner-like advertisements. Next, we discard any image smaller than 100 pixels due to their inadequate detail for analysis. We also exclude images containing specific keywords such as ``logo'' and ``social,'' which typically denote non-relevant graphics or branding content.

To guarantee the high quality of images included in our benchmark, we first employed CLIP similarity \citep{hessel-etal-2021-clipscore} to rank the remaining images for cultural relevance. Based on a manual annotation of images for 200 CCSK assertions, to assess their relevance to the CCSK, we set a threshold of 23 to ensure culturally relevant images (precision = 0.92, recall = 0.96). Images below this score were discarded. Higher-scoring images were more likely to be selected for question creation.

\subsection{Stereotypes and Biases}
\label{app:stereotypes}
To ascertain the representational fairness of our dataset, we implemented a Sentence-Level Stereotype Classifier,\footnote{\url{https://huggingface.co/wu981526092/Sentence-Level-Stereotype-Detector}} a transformer-based model, for detecting stereotypical content within the dataset's questions. This model's efficacy in classifying sentences based on the presence of stereotypes or anti-stereotypes was evaluated across various dimensions including race, gender, religion, and profession. The classifier identified relatively few stereotypical instances: 69 cases pertained to race, 44 to gender, 22 to religion, and 8 to profession. In contrast, anti-stereotypical content was more prevalent, with 169 cases for race, 25 for religion, 23 for gender, and 7 for profession. A significant portion of the data, 923 instances, did not correlate with any stereotypical or anti-stereotypical categories, underscoring the minimal presence of biased content in the dataset. These findings support the dataset's utility in facilitating unbiased and culturally comprehensive studies.

\subsection{VLMs Used for Benchmarking}
\label{app:models}
We benchmark the following state-of-the-art open-source VLMs on our proposed \dataset dataset: \textsc{Blip2}~\cite{li2023blip}, \textsc{InstructBlip}~\cite{dai2024instructblip},
\textsc{mBLIP}~\cite{geigle2023mblip}
\textsc{PalliGemma}~\cite{beyer2024paligemmaversatile3bvlm}
\textsc{Llava1.5}~\cite{liu2023llava}, \textsc{Llava\_Next}~\cite{liu2024llavanext}, \textsc{Idefics2}~\cite{laurenccon2024matters}, and \textsc{Intern-VL 1.5}~\cite{chen2024far}. These models were selected based on their release year and parameter size (3 to 25 billion) to test how these aspects affect cultural understanding. \textsc{InstructBlip}, fine-tuned with instruction tuning, is compared to \textsc{Blip2} to see if instruction tuning enhances cultural understanding. \textsc{Idefics2}, with 8 billion parameters, is evaluated for its performance on open datasets, surpassing larger models. \textsc{Intern-VL 1.5}, with 25 billion parameters, bridges the gap between open-source and proprietary models, showing strong multimodal benchmark performance, even outperforming proprietary models on some benchmarks. For each model, we use the default text-generation parameters as found in their HuggingFace code repository which include a greedy decoding strategy with the temperature set to 1. Finally, we also evaluate closed-source models -- \textsc{GPT-4} (GPT-4o), \textsc{Gemini} (Gemini-Pro-Vision 1.0) and \textsc{Claude} (Claude 3.5 \citep{anthropic2024claude}) -- using their API endpoints.

\begin{figure}[ht!]
\centering
\includegraphics[width=\columnwidth]{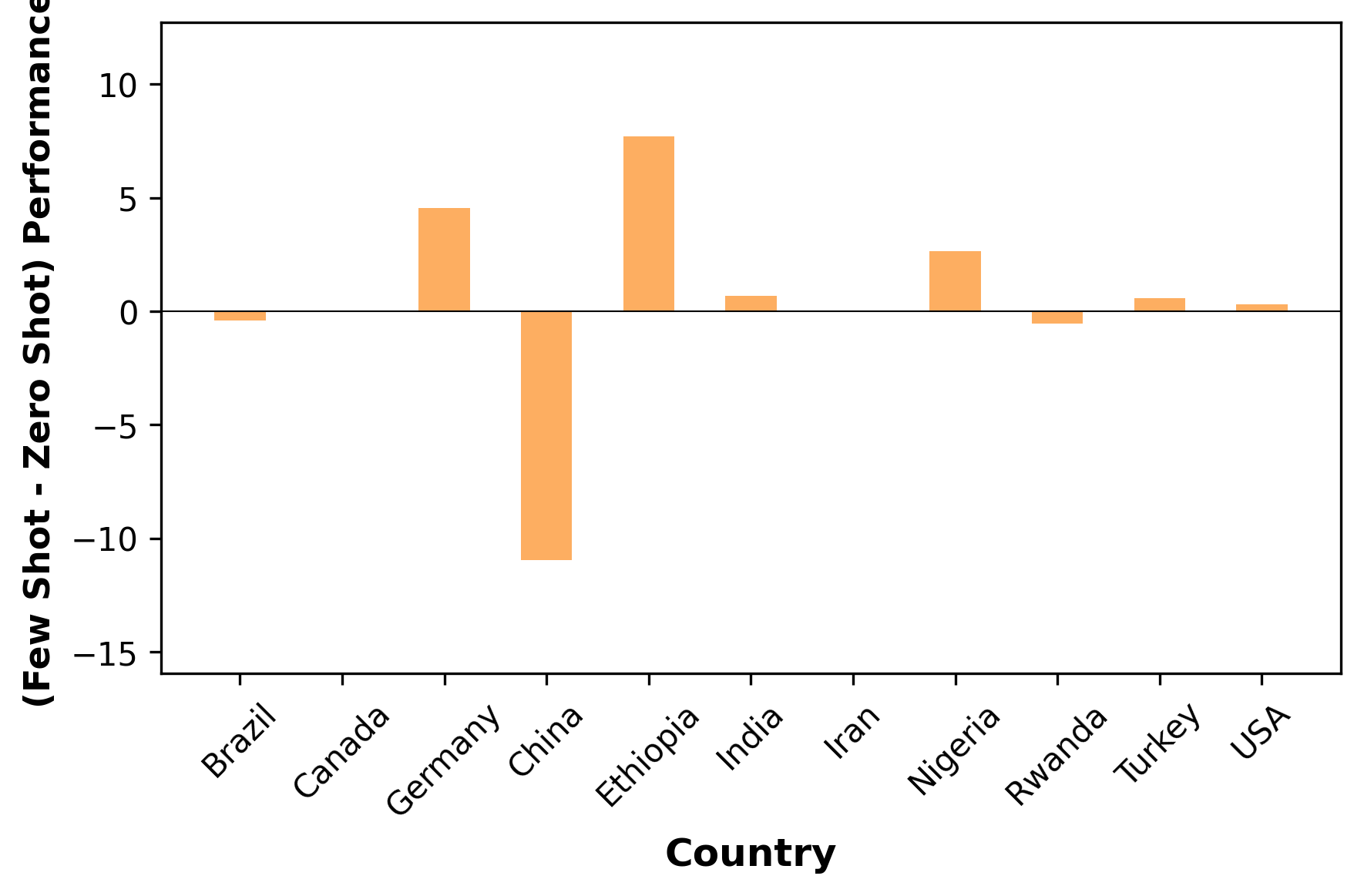}
\caption{Delta graph for the change in performance from zero-shot to few-shot prompting using \textsc{GPT-4}.}
\label{fig:gpt4-delta}
\vspace{-10pt}
\end{figure}

\subsection{Few-Shot Evaluation of GPT-4}
\label{app:few_shot}
We conduct a \textbf{few-shot} evaluation of GPT-4 (best performing model) to determine whether the \dataset benchmark can be solved by guiding the models with a few examples. In this setup, we include one example per country (11 examples total). The few-shot prompt is detailed in \Cref{app:fs_inf}. Our analysis (\Cref{fig:gpt4-delta}) reveals that few-shot prompting does not consistently improve performance over zero-shot, despite examples from all countries. While some countries like Germany, Ethiopia, and Nigeria showed improvements (3-8\%), others such as Brazil, China, India, and Rwanda experienced performance drops or minimal gains. This suggests that few-shot prompting may not be uniformly beneficial across cultural contexts and that \textsc{GPT-4}'s performance on culturally nuanced tasks largely depends on its pre-existing knowledge. These results highlight the challenging nature of \dataset and indicates the need for more advanced methods to enhance model performance on cultural understanding tasks.

\begin{table*}[]
\begin{tabular}{lll} \toprule
\textbf{Facet}      & \textbf{Qustion/Answer input to GPT-}4                                                                                           & \textbf{Classified subcategory} \\ \midrule
Food       & \begin{tabular}[c]{@{}l@{}}Q: What is the traditional name of the bread in the picture?\\ A: Barbari\end{tabular}       & Type/Name              \\
Clothing   & \begin{tabular}[c]{@{}l@{}}Q: In which Indian state is this dressing style most popular?\\ A: Punjab\end{tabular}       & Location/Region        \\
Drink      & \begin{tabular}[c]{@{}l@{}}Q: What is the taste of the pictured alcoholic beverage\\ A: Cinnamon\end{tabular}           & Taste                  \\
Rituals    & \begin{tabular}[c]{@{}l@{}}Q: In Nigerian culture, what does the image represent?\\ A: spiritual activity\end{tabular}  & Beliefs and Customs    \\
Traditions & \begin{tabular}[c]{@{}l@{}}Q: What is the name of the national anthem related to this flag?\\ A: Oh Canada\end{tabular} & Music/Instruments     \\ \bottomrule
\end{tabular}
\caption{
Examples of subcategories assigned by GPT-4 for different question-answer pairs from each facet. "Q" represents the question, and "A" represents the answer in the "Qustion/Answer input to GPT-4 " column.}
\label{tab:examples}
\end{table*}

\begin{figure*}[!ht]
    \centering
    \begin{subfigure}[b]{0.49\textwidth}
        \centering
        \includegraphics[width=\textwidth]{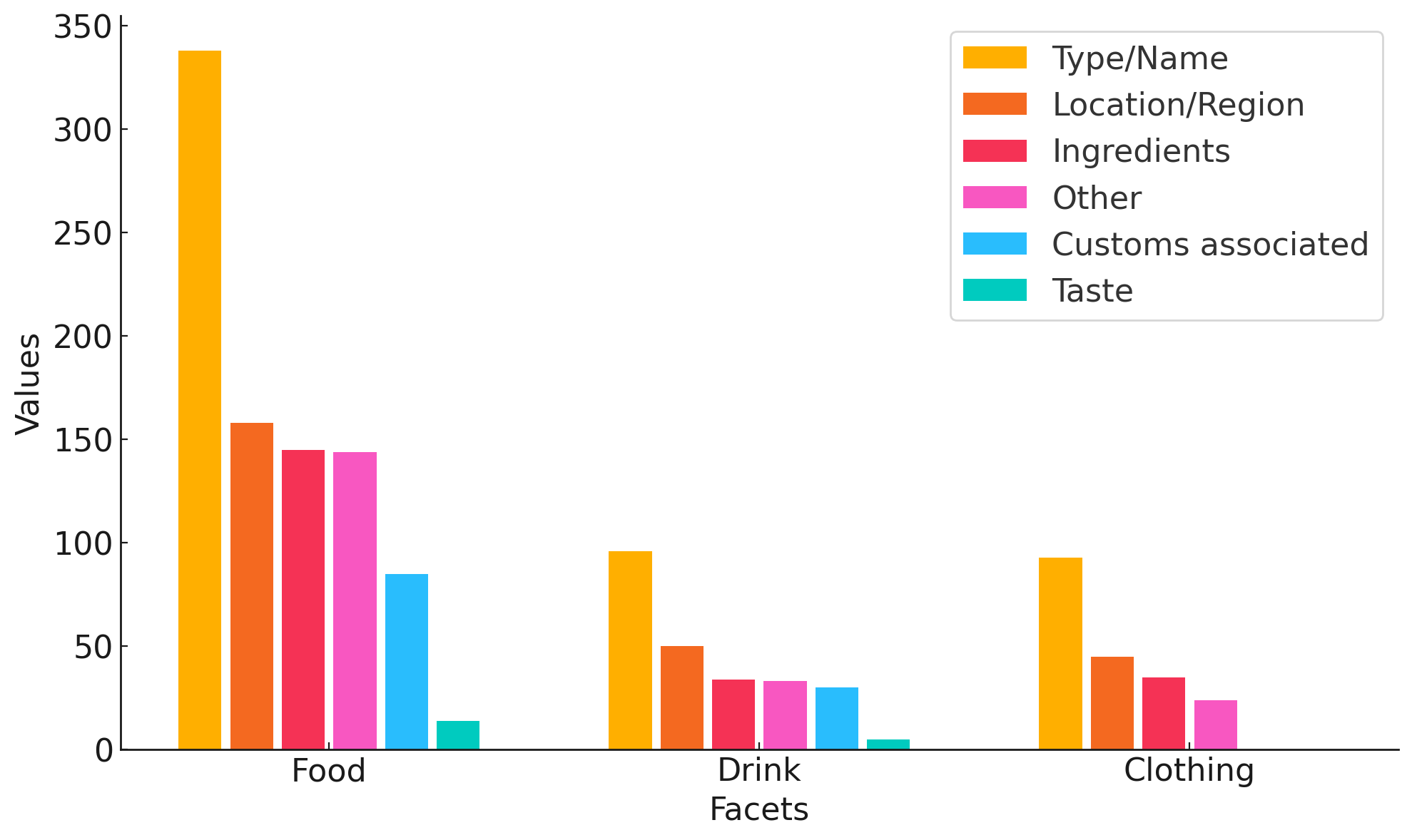}
        \label{fig:food_drink_clothing}
    \end{subfigure}
    \begin{subfigure}[b]{0.49\textwidth}
        \centering
        \includegraphics[width=\textwidth]{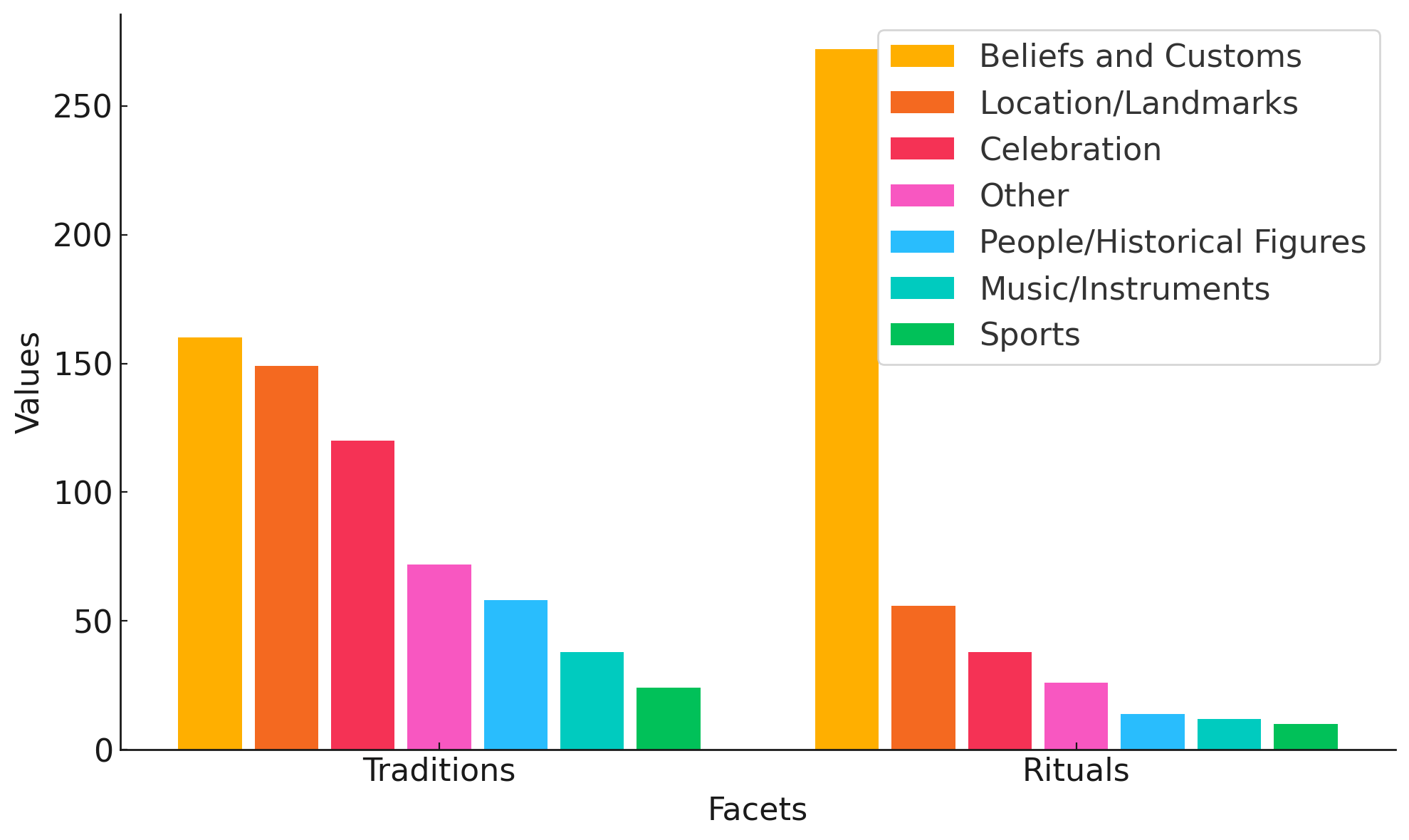}
        \label{fig:traditions_rituals}
    \end{subfigure}
    \vspace{-20pt}
    \caption{Breakdown of facets into various subcategories. The plot on the left illustrates the detailed subcategories for the Food, Drink, and Clothing facets, while the plot on the right presents the corresponding breakdown for the Rituals and Traditions facets.}
    \label{fig:combined_facets}
\end{figure*}

\begin{table*}[!ht]
    \centering
    \begin{adjustbox}{max width=\linewidth}
    \begin{tabular}{lccccccc}
        \toprule
        \multirow{2}{*}{Fine-grained Categories} & \multicolumn{2}{c}{Food} & \multicolumn{2}{c}{Drink} & \multicolumn{2}{c}{Clothing} & \multirow{2}{*}{Average} \\
        \cmidrule(r){2-3} \cmidrule(r){4-5} \cmidrule(r){6-7}
        & GPT-4V & InternVL & GPT-4V & InternVL & GPT-4V & InternVL & \\
        \midrule
        Type/Name           & 63.1 & 41.9 & 71.4 & 54.9 & 60.2 & 51.6 & 57.18 \\
        Location/Region     & 65.6 & 43.9 & 48.9 & 33.3 & 77.3 & 72.7 & 56.95 \\
        Customs Associated  & 59.2 & 51.2 & 57.1 & 42.8 & 40.0 & 57.7 & 51.33 \\
        Ingredients         & 54.7 & 45.9 & 78.1 & 59.3 & N/A  & N/A  & 59.50 \\
        Taste               & 42.8 & 35.7 & 20.0 & 20.0 & N/A  & N/A  & 29.62 \\
        Other               & 59.1 & 63.5 & 46.8 & 43.7 & 52.9 & 50.0 & 52.66 \\
        \bottomrule
    \end{tabular}
    \end{adjustbox}
    \caption{Performance of GPT-4V and InternVL on subcategories of Food, Drink, and Clothing Facets}
    \label{tab:food_drink_clothing}
\end{table*}

\begin{table*}[!ht]
    \centering
    \begin{adjustbox}{max width=\linewidth}
    \begin{tabular}{lccccc}
        \toprule
        \multirow{2}{*}{Fine-grained Categories} & \multicolumn{2}{c}{Traditions} & \multicolumn{2}{c}{Rituals} & \multirow{2}{*}{Average} \\
        \cmidrule(r){2-3} \cmidrule(r){4-5}
        & GPT-4V & InternVL & GPT-4V & InternVL & \\
        \midrule
        Beliefs and Customs     & 57.1 & 35.9 & 58.9 & 39.7 & 47.9 \\
        Location/Landmarks      & 59.6 & 45.2 & 71.1 & 57.9 & 58.4 \\
        Celebration             & 81.6 & 78.3 & 86.8 & 64.2 & 77.7 \\
        Music/Instruments       & 50.0 & 28.9 & 57.1 & 57.1 & 48.3 \\
        Sports                  & 73.9 & 73.9 & 58.3 & 58.3 & 66.10 \\
        People and Hist. Figures & 61.4 & 50.9 & 56.0 & 48.0 & 54.1 \\
        Other                   & 69.6 & 57.9 & 50.0 & 60.0 & 59.4 \\
        \bottomrule
    \end{tabular}
    \end{adjustbox}
    \caption{Performance of GPT-4V and InternVL on subcategories of Traditions and Rituals Facets}
    \label{tab:traditions_rituals}
\end{table*}

\begin{figure}[!ht]
    \centering
    \includegraphics[width=\linewidth]{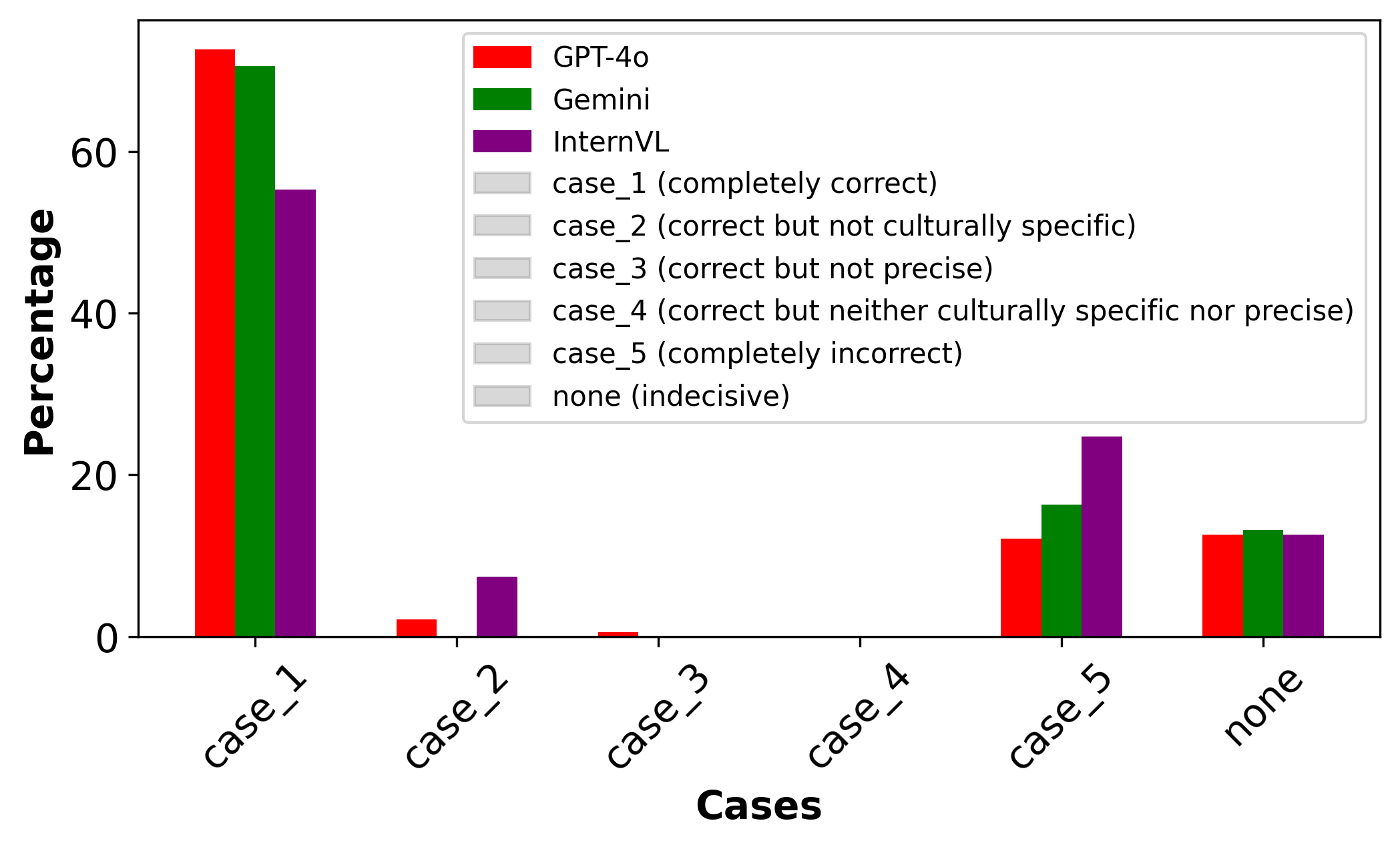}
    \caption{Distribution of human judgments for model answers in India across different models (\textsc{GPT-4o}, \textsc{Gemini}, \textsc{Intern-VL}). \textsc{GPT-4o} and \textsc{Gemini} show the highest percentage of completely correct answers (case\_1), while \textsc{Intern-VL} has a significant percentage of completely incorrect answers (case\_5).}
    \label{fig:hum_jud}
    \vspace{-10pt}
\end{figure}

\begin{figure}[!ht]
    \centering
    \includegraphics[width=\linewidth]{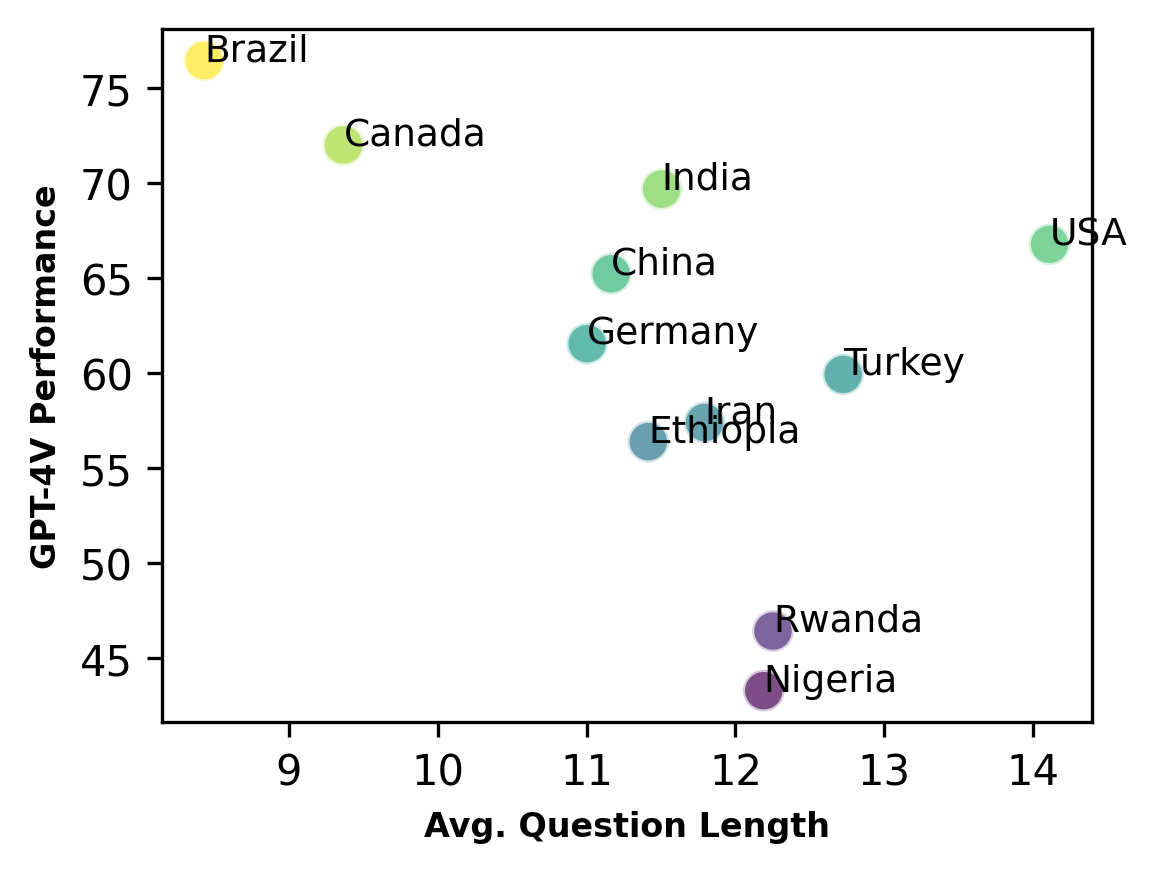}
    \caption{Scatter plot of GPT-4V performance versus average question length across different countries}
    \label{fig:corelation}
    \vspace{-10pt}
\end{figure}

\subsection{Analysis of Performance Across Cultural Facets}
\label{app:facets}

To better understand the performance disparities between the different facets, we categorise the image-question-answer triplets in our dataset into more fine-grained categories based on the aspect of the facet being probed in the question. 
More specifically, the sub-categories include \textit{Type / Name}, \textit{Location / Region}, \textit{Customs associated}, \textit{Ingredients}, \textit{Taste}, \textit{Other} (for food, clothing, and drinks facets), and \textit{Beliefs and Customs}, \textit{Location / Landmarks}, \textit{Celebration}, \textit{Music / Instruments}, \textit{Sports}, \textit{People / Historical Figures}, \textit{Other} (for traditions and rituals facets).
These fine-grained categories are inspired by the categorization in MaRVL \citep{liu-etal-2021-visually} for the traditions and rituals facets and FoodieQA \citep{li2024foodieqamultimodaldatasetfinegrained} for the food, drink, and clothing facets. We prompt GPT-4 with question, answer, and the original facet along with a list of fine-grained categories and several in-context examples to perform this categorization. A few examples from this exercise are shown in \Cref{tab:examples}. We report the number of image-question-answer triplets belonging to each fine-grained category in \Cref{fig:combined_facets}.

While the most popular fine-grained category for the food, drink, and clothing facets corresponds to identifying the type or name of the entity, a significant proportion of the samples (61.8\% for Food, 61.3\% for Drink, 52.7\% for Clothing) require more detailed knowledge such as associated customs. The samples from traditions and rituals require more diverse knowledge, with the sub-categories of \textit{Beliefs and Customs}, \textit{Location / Landmarks}, and \textit{Celebration} being the most prevalent.

We summarize the results obtained for different subcategories for GPT-4 and InternVL in \Cref{tab:food_drink_clothing} and \Cref{tab:traditions_rituals}. From these results, we observe that among the food, clothing, and drink facets, on average, models tend to perform better on questions that involve identifying the name, location, and ingredients (only applicable to food and drink facets) of the concept. However, they perform relatively poorly on questions probing associated customs and taste (only applicable to food and drink facets). Similarly, for the rituals and traditions facets, models show strong performance in identifying celebrations, locations, landmarks, and sports, but perform relatively poorly on identifying beliefs, customs and music / instruments.

\paragraph{Why do certain models perform better on specific cultural facets?} We analyze three factors that could lead to disparity in a model's performance across facets. From \Cref{tab:food_drink_clothing} and \Cref{tab:traditions_rituals}, we observe that there are stark differences in performance across different fine-grained categories. For instance, the relatively better performance of GPT-4 on questions about traditions and beliefs can be attributed to a couple of fine-grained categories, such as celebrations (Q: “For which holiday season are these items in the image popular?”, A: “Christmas”), landmarks (Q: “What is the name of this famous Hindu temple shown above?”, A: “Janaki temple”), and sports (Q: “What are the people in the picture practicing?”, A: “Wushu”).

Another source of disparity in the performance could be the disparity in inherent difficulty levels of questions belonging to each facet. To investigate this, we calculate human performance for each facet and observe minimal differences (performance for food - 74\%, clothing - 72.7\%, drinks - 74.5\%, rituals - 71.1\%, traditions - 73.7\%), suggesting that this is unlikely to be the case.

Finally, we investigate if the disparity in model performance across facets is correlated with the representation of each facet in the model’s pre-training data. We conduct this study for the best-performing open-source model – InternVL. We randomly sample 1.3 million data points from LAION (the pretraining data for Intern-VL) and check how many samples in the pretraining data contain at least one answer string from our benchmark corresponding to a given facet. Once we get the counts, we normalize them by the total number of answers within each facet, since facets with more number of answers will naturally have more matches in the pretraining data. The relative percentages for each facet are as follows: Food (46.6\%), Clothing (6.5\%), Drink (7.8\%), Rituals (25.1\%), Traditions (13.8\%), and Others (0.2\%).
We observe that the food facet has the highest representation, followed by rituals and traditions. However, this does not align with the performance trends observed for Intern-VL, where the highest performance is seen for clothing, followed by traditions, rituals, food, and drink. This suggests that factors beyond the occurrence of concepts in the pre-training data contribute to the disparity in model performance across different facets. Understanding these factors presents an intriguing area for future research.

\subsection{Qualitative Analysis of Human Performance}
\label{app:human_performance}
We qualitatively investigate why countries like Nigeria and Rwanda exhibit relatively lower human performance. We identify two major contributing factors. First, we have used country as a proxy for a cultural group, which might be particularly inaccurate for these countries. There may be subcultures within these countries where the same concept holds different meanings, leading to varied interpretations. This is especially relevant for visually similar items. For example, for the question: ``What's the item that the people are beating called in the local parlance?,'' the answers received are \textit{\`Il\`u}, \textit{Igba}, and \textit{drums}. The first two are also types of drums, with the former used in Yoruba culture and the latter in Igbo culture. Depending on the respondents' cultural background, their answers may have varied. Secondly, we also found that annotators often disagreed on questions that required identifying geographical locations. For example, for the question: ``What part of Rwanda are the crops shown in the image grown more?'' the answers are \textit{Gisagara}, \textit{Gicumbi District}, and \textit{Nyamagabe}. These types of questions, especially for Rwanda, might have contributed to the lower performance

\subsection{Human Judgment of Model Predictions}
\label{app:human-judgment}

\begin{figure*}
    \centering
    \includegraphics[width=\linewidth]{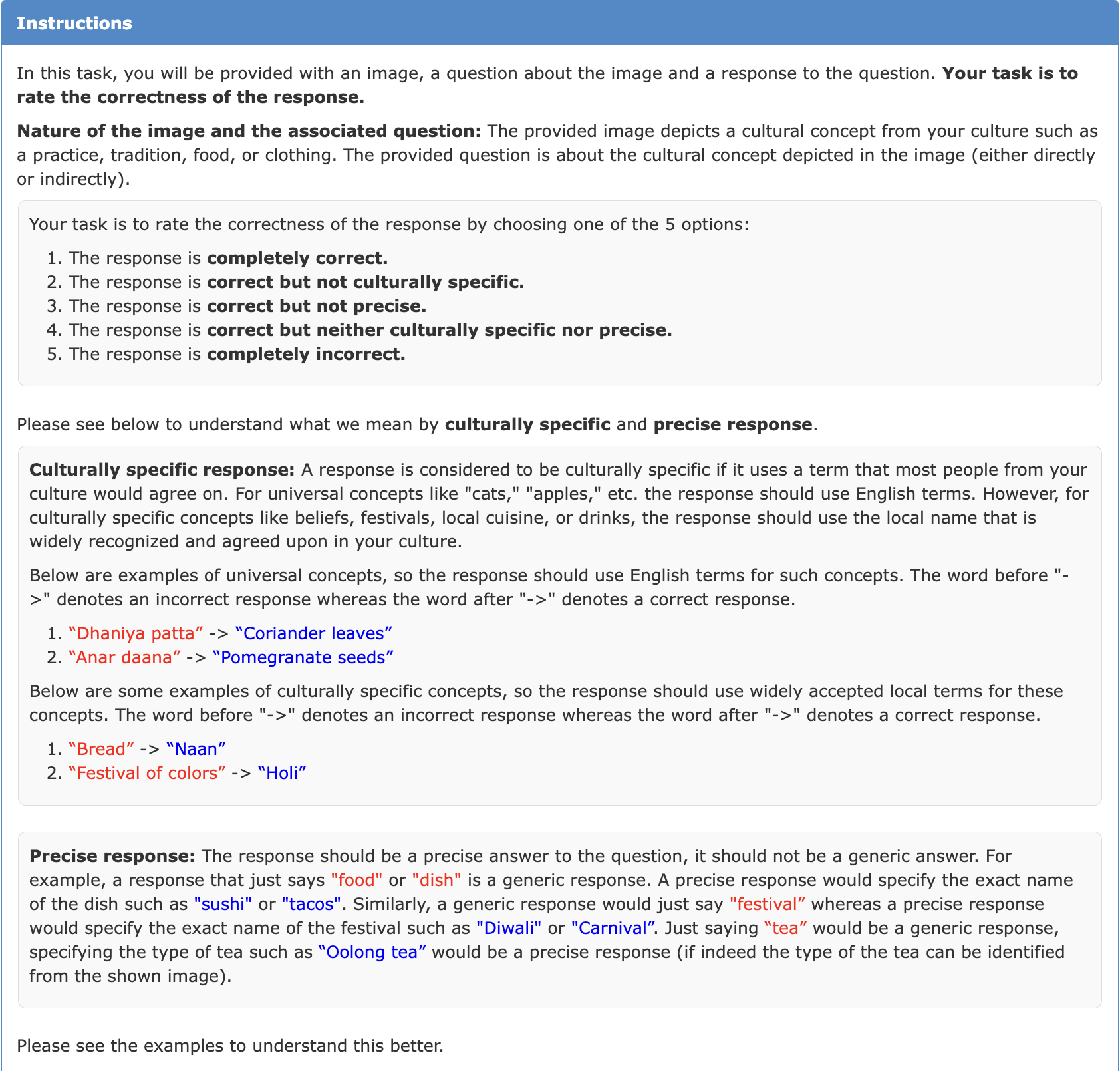}
    \caption{The instructions given to annotators to evaluate answers generated by various models. To assist with writing, we provide clear guidelines and offer multiple examples showcasing both good and poor practices.}
    \label{fig:mturk_evaluation}
\end{figure*}

We perform human evaluation of model responses for questions from India. Five human annotators rate each answer on a scale of 1 to 5: 1 (completely correct), 2 (correct but not culturally specific), 3 (correct but not precise), 4 (correct but neither culturally specific nor precise), and 5 (completely incorrect). The instructions given to the annotators can be found in \Cref{fig:mturk_evaluation}.

\subsection{Behind the scenes: Journey of how \dataset came into place}
\label{app:challenges}
The journey of creating the CulturalVQA dataset was shaped by various design decisions, challenges, and lessons learned. This section aims to outline our motivations, initial ideas, and the obstacles we encountered, with the hope of guiding others who are interested in building similar datasets.

\paragraph{Motivation and Initial Idea} The project was primarily motivated by the lack of comprehensive benchmarks to evaluate cultural understanding in vision-language models (VLMs) across a broad set of countries. We wanted to create a resource that would holistically test these models' cultural knowledge. We were looking into a source for obtaining culturally diverse images. The initial spark for the dataset came from the CCSK \citep{candle2023} and MMC4 papers \citep{mmc4}, inspiring exploration into leveraging the images in the C4 corpus \citep{10.5555/3455716.3455856}.

\paragraph{Early Efforts and Challenges}
In December 2023 and January 2024, we focused on scraping, filtering, and conducting quality analysis on the images from the C4 corpus filtered using cultural commonsense knowledge assertions from \textsc{CANDLE}. Initially, our goal was to create a large-scale dataset semi-automatically, covering about 100 countries. We wanted to leverage LLM-based question generation methods to achieve this. By March 2024, we had built an early version of \dataset that included 12 countries. We used GPT-4 to generate cultural questions based on the CCSK information and metadata like captions, object information and entity tags from Google Lens \citep{googlelensapi}. However, we soon found several issues with this dataset. For instance, GPT-4 performed exceptionally well on the dataset achieving results above 90\% for countries like India, Germany, and Poland. The open-source models like LLaVA-Next \citep{liu2024llavanext} were not very far behind. These results echo the observations by \citet{baek2024evaluatingvisualculturalinterpretation}, who build a dataset using a similar method for Korean culture and observe that models like GPT-4 and Gemini surpass human performance on their dataset. On further analysing the questions, we found that they required only a coarse-grained understanding of visual content and did not adequately probe for cultural nuance. This highlighted the limitations of building such geo-diverse and cultural datasets using existing LLMs. Hence, we reevaluated our approach, and we decided to involve human annotators to enhance cultural depth and authenticity.

\paragraph{Note on Filtering Images} We aimed to use automated methods to create an image corpus for building the \dataset benchmark. This idea originated from the need for a large-scale dataset, which would be impractical to gather solely through human efforts. The internet, as a vast and diverse source of imagery, provided an opportunity to build a culturally rich image corpus. However, since we decided to involve human evaluators, our final approach was not entirely automated; it incorporated human input to further refine the dataset. This human refinement led to the removal of 19.64\% of the images, highlighting that automated methods alone are still insufficient for constructing such high-quality datasets. Future work could explore methods to bridge this gap.

Even though we obtain a culturally relevant corpus from our image selection method, leveraging only the English portion of Common Crawl has its limitations, as it predominantly contains popular concepts from well-represented cultures. We hypothesize that utilizing the multilingual segments of Common Crawl could help uncover more rare cultural concepts and corresponding images, leading to a more diverse and inclusive dataset.

\paragraph{Annotator Selection and Pilot Studies} We then explored crowdsourcing platforms and ultimately chose Amazon Mechanical Turk (MTurk) due to its easy-to-use interface. We also considered Prolific, but its lack of interface customization led us back to MTurk. Our initial pilot began with India where we spent about a month conducting pilots to debate and fine-tune the guidelines. We believe this is a very important step to collect high-quality data and it is worth spending a lot of time on this. Once we were satisfied by the guidelines we aimed for larger-scale annotation for multiple countries. Unfortunately, we quickly discovered a major challenge: MTurk had almost no active annotators for countries outside the US, Canada, India, and Brazil. We tried to collect data from the Philippines, Indonesia, Japan, Germany, France, China, Iran, and Morocco, but found almost no willing annotators. This taught  us the difficulty of recruiting diverse annotators through traditional platforms. This is also an important bottleneck for building representative datasets required to build inclusive models.

\paragraph{Shifting to Community Involvement} 
To address the limitations of cultural representation, we turned to more diverse communities by partnering with Mila and Masakhane for annotations. We conducted several workshops and maintained ongoing communication with annotators through extensive email threads to provide consistent feedback. However, we faced challenges with providing timely feedback to MTurk participants compared to our direct community engagements, which resulted in discarding a significant amount of data from MTurk due to poor adherence to guidelines.

Managing a large group of annotators across different time zones added further complexity, emphasizing the need for scalable platforms or outsourcing to enhance efficiency. After completing the paper, we discovered platforms like CloudConnect, which have been used in works such as \citep{bhatia2024localconceptsuniversalsevaluating} to collect data from a larger number of countries. However, they also faced similar challenges in obtaining high-quality data, with poor communication with annotators leading to the rejection of numerous data points. This highlights the common struggle of balancing scale and quality in annotation processes across diverse regions.

\paragraph{Key Takeaways} Building the CulturalVQA dataset was a challenging yet rewarding journey. What began as an automated, LLM-driven approach evolved into one deeply rooted in human annotation. Our biggest takeaway is that human input remains irreplaceable in creating culturally rich datasets—at least for now. Additionally, leveraging a scalable platform with a dedicated, diverse pool of annotators, combined with effective and timely communication, is essential for achieving high-quality results. Choosing the right annotators is critical, as their contributions directly impact the dataset’s quality. Conducting multiple pilot studies was invaluable in helping us identify the best annotators and refine our process.

By sharing our experiences—from initial ideas to refining our annotation methods—we hope to provide guidance to others facing similar challenges in creating culturally diverse benchmarks for VLMs. We believe that our journey offers useful insights for building more inclusive, high-quality datasets in the future.

\subsection{Prompt for Few-Shot Inference using GPT-4}
\label{app:fs_inf}
\begin{tcolorbox}[colback=gray!5!white, 
                  colframe=gray!75!black, 
                  title=Prompt used for few-shot inference, 
                  fonttitle=\bfseries, 
                  sharp corners=all] 
You will be given an image depicting a cultural concept and a question about the image. Answer the question with a precise, culturally specific response (e.g., `sushi' instead of `food', `Diwali' instead of `festival') of 1-3 words. Here are some examples of the described task.

\{image\}

\{question\}

\{answer\}
\end{tcolorbox}

\subsection{Prompt for VLM Inference}
\label{app:vl_inf}

\begin{tcolorbox}[colback=gray!5!white, 
                  colframe=gray!75!black, 
                  title=Prompt used to test VLM inference, 
                  fonttitle=\bfseries, 
                  sharp corners=all] 
You will be given an image depicting a cultural concept and a question about the image. Answer the question with a precise, culturally specific response (e.g., `sushi' instead of `food', `Diwali' instead of `festival') of 1-3 words.
\end{tcolorbox}

\subsection{System Prompt for the Evaluation Metric}
\label{app:lave}

\begin{tcolorbox}[colback=gray!5!white, 
                  colframe=gray!75!black, 
                  title=System prompt used for the LAVE evaluation metric, 
                  fonttitle=\bfseries, 
                  sharp corners=all] 
You are an expert cultural anthropologist tasked with evaluating the correctness of candidate answers for cultural visual question answering. Given a question, a set of reference answers by an expert, and a candidate answer by a model, please rate the candidate answer's correctness. Use a scale of 1-2, where 1 indicates an incorrect, irrelevant, or imprecise answer, and 2 indicates a correct and precise answer. Specify the rating in the format `rating=X', where X is either 1 or 2. Also, provide the rationale for your rating. 
\end{tcolorbox}

\subsection{Inference Using Closed-Source Models}
\label{app:inf_closed}

In this section, we provide the sample code used for accessing \textsc{Gemini} and \textsc{GPT-4}.

For performing inference using \textsc{Gemini}, we leverage the Vertex AI API for \textsc{Gemini} with multimodal prompts. The code snippet for inference is provided below.

\begin{lstlisting}[language=Python, caption=Code snippet for accessing Gemini using API]
import google.generativeai as genai

genai.configure(api_key=<api_key>)
model = genai.GenerativeModel('gemini-pro-vision')

response = model.generate_content([question, image], 
stream=False, 
request_options={"timeout": 600})
response.resolve()
predicted_answer = [response.text]
\end{lstlisting}

\begin{figure*}
    \centering
    \includegraphics[width=\linewidth]{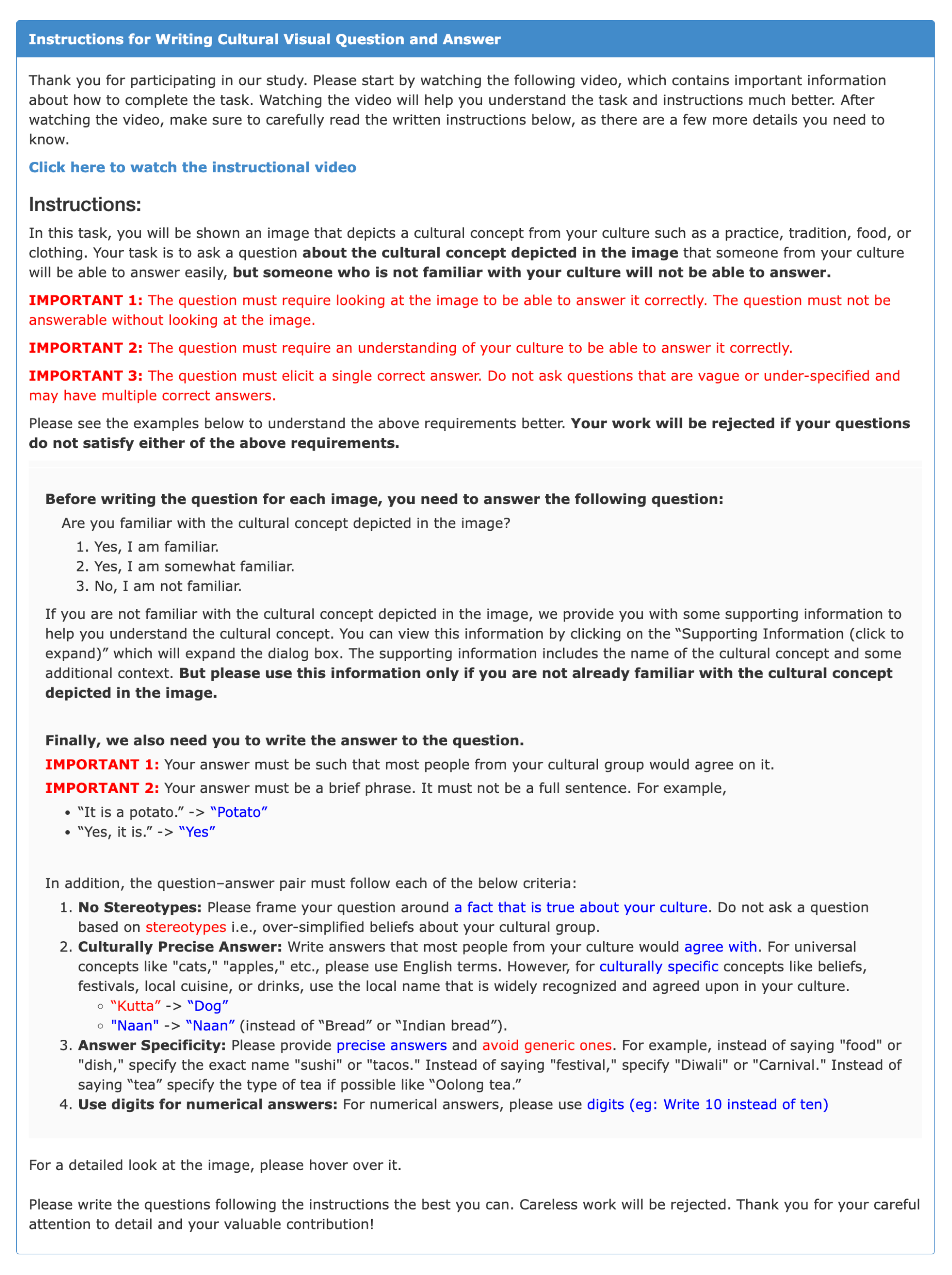}
    \caption{Instructions given to annotators from India to write questions and answers for images. Similar instructions, with different examples, were given to annotators from other countries. To assist with writing, we provide a brief video detailing our task and guidelines, along with multiple examples showcasing both good and poor practices (examples not included here).}
    \label{fig:mturk_ques}
\end{figure*}

\subsection{Instructions for Human Question Generation}
\label{app:instructions-quest}
We iteratively refined the guidelines provided to human annotators, conducting multiple pilot studies on MTurk to fine-tune these guidelines until we obtained satisfactory quality in the questions from the annotators. The detailed instructions given to the annotators for writing questions can be found in \Cref{fig:mturk_ques}.

\begin{figure*}
    \centering
    \includegraphics[width=\linewidth]{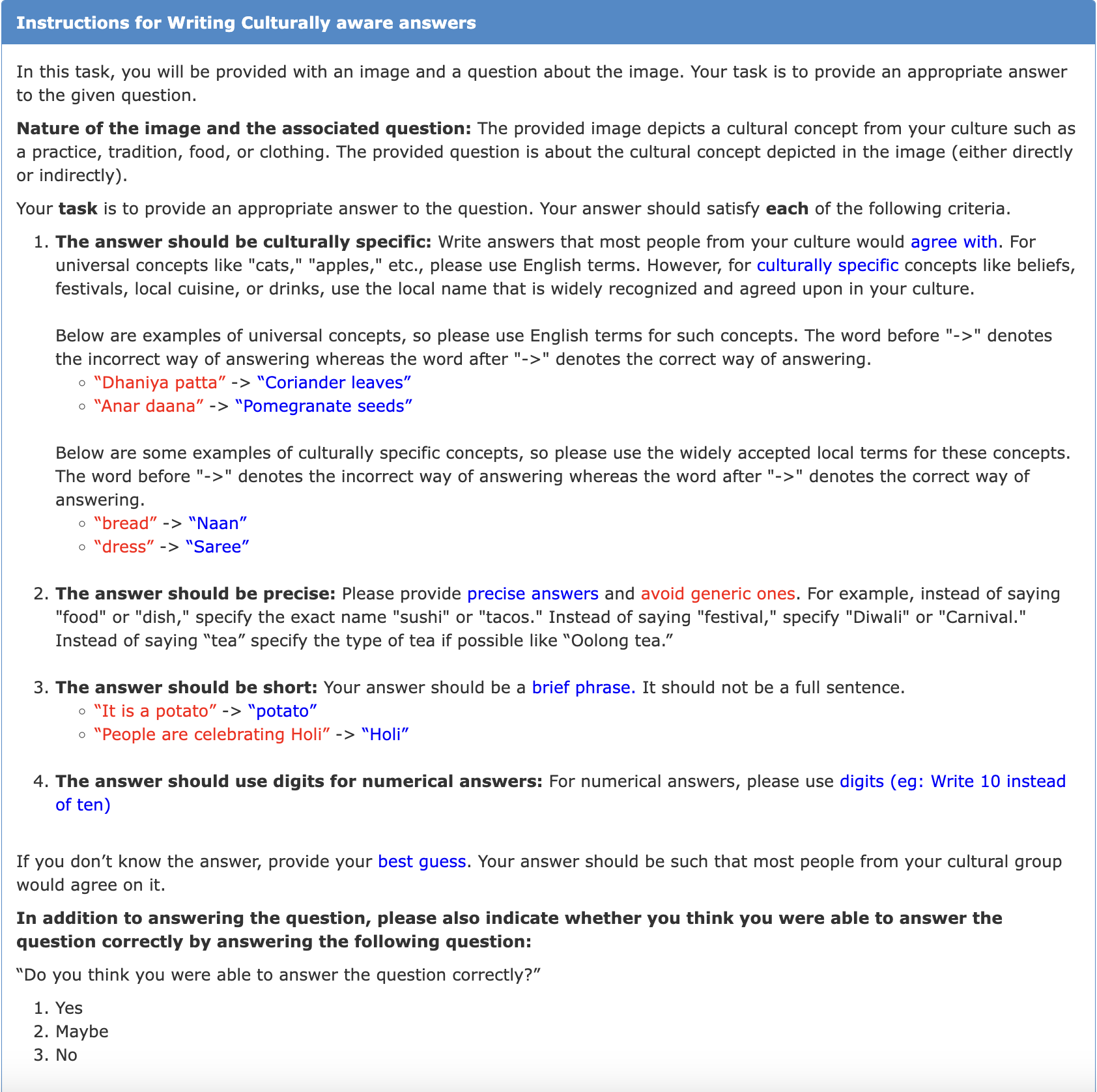}
    \caption{The instructions given to annotators from India to write answers for questions collected for images. Similar instructions, with different examples, were given to annotators from other countries. To assist with writing, we provide clear guidelines and offer multiple examples showcasing both good and poor practices.}
    \label{fig:mturk_ans}
\end{figure*}

\subsection{Instructions for Human Answer Generation}
\label{app:instructions-answer}
Similar to the question generation guidelines, we conducted multiple pilot studies on MTurk to refine the instructions, ensuring that annotators adhered to the criteria required for writing answers. The instructions provided to the annotators for collecting answers are detailed in \Cref{fig:mturk_ans}.

\end{document}